\documentclass{article}

\usepackage[final]{mcap_style}

\usepackage[utf8]{inputenc}
\usepackage[T1]{fontenc}
\usepackage{hyperref}
\usepackage{url}
\usepackage{booktabs}
\usepackage{amsfonts}
\usepackage{amsmath}
\usepackage{amssymb}
\usepackage{nicefrac}
\usepackage{microtype}
\usepackage{xcolor}
\usepackage{graphicx}
\usepackage{subcaption}
\usepackage{algorithm}
\usepackage{algorithmic}
\usepackage{multirow}
\usepackage{colortbl}
\usepackage{tikz}
\usetikzlibrary{positioning, arrows.meta, fit, calc, decorations.pathreplacing, shapes.geometric}
\usepackage{pgfplots}
\pgfplotsset{compat=1.18}
\usepackage{listings}
\usepackage{enumitem}
\usepackage{wrapfig}
\usepackage{tabularx}
\usepackage{needspace}
\usepackage[section]{placeins}
\usepackage{float}

\definecolor{nveblue}{HTML}{377EB8}
\definecolor{nvered}{HTML}{E41A1C}
\definecolor{nvegreen}{HTML}{4DAF4A}
\definecolor{nveorange}{HTML}{FF7F00}
\definecolor{nvepurple}{HTML}{984EA3}
\definecolor{nvegray}{HTML}{999999}
\definecolor{tier0}{HTML}{2166AC}
\definecolor{tier1}{HTML}{67A9CF}
\definecolor{tier2}{HTML}{D1E5F0}
\definecolor{w4a8color}{HTML}{4DAF4A}
\definecolor{w4a16color}{HTML}{377EB8}
\definecolor{llamacppcolor}{HTML}{E41A1C}

\title{MCAP: Deployment-Time Layer Profiling for \\
Memory-Constrained LLM Inference}

\author{%
  Anurita Das \\
  Genovation Technological Solutions Pvt Ltd \\
  \texttt{anurita@genovationsolutions.com} \\
  \texttt{\url{https://github.com/genovationtech/nve}} \\
}

\begin{document}

\maketitle

\begin{abstract}
Deploying large language models to heterogeneous hardware is constrained by
memory, not compute: a 4-bit 8B model still requires $\sim$5--6\,GB of
resident VRAM, ruling out most consumer GPUs, mobile accelerators, and
older cloud silicon. Existing post-training quantization fixes the
precision choice at calibration time, so the set of devices on which a
given quantized model is viable is decided before it ever reaches
hardware. We introduce a \emph{load-time control layer} for LLM inference: the
per-layer precision decision is made on the target device from a
lightweight runtime signal rather than fixed offline, which substantially
widens the set of reachable operating points for a given set of weights.

We introduce \textbf{MCAP} (Monte Carlo Activation Profiling), a
load-time layer importance estimator: a 60-second gradient-free
profiler that produces a per-layer importance signal from 12
calibration prompts, and \textbf{NVE}, a Rust+CUDA
inference engine that uses that signal to drive two coupled per-layer
decisions: precision dispatch (W4A8 vs.\ W4A16) and residency tier
(GPU / RAM / SSD). A profile is a 338-byte JSON keyed by architecture
rather than a new set of weights, and each target computes its own on
load. This couples two decisions that prior systems handle separately:
precision routing and memory placement, from a single runtime signal.

NVE delivers 1.5--1.8$\times$ higher decode throughput than llama.cpp
Q4\_0 on NVIDIA T4 across 1B, 3B, and 8B Llama models, with the
advantage over W4A16 widening from 2.3$\times$ at 1B to 2.9$\times$ at
8B. More importantly, the same signal drives a three-tier weight pager
that reaches operating points conventional GPU-resident inference
cannot: Llama-3.2-3B ($\sim$6.4\,GB BF16) runs in 2\,GB at 2.31 tok/s,
and Llama-3.1-8B ($\sim$16\,GB BF16) runs in 4\,GB at 1.25 tok/s, both
with no observable degradation on the evaluated tasks relative to the
unconstrained baseline. Against AutoAWQ at matched
4-bit precision, MCAP is within binomial CI on HellaSwag (54\% vs.\
52\%) with no offline calibration and no weight modification. A single
338-byte profile, SHA-256-identical across six parallel containers on
two NVIDIA silicon generations (Turing T4 and Ampere A10G), drives
correct 87.5\% task accuracy on all 12 runs at 0\,ms on-device
profiling cost. We report where the method breaks (aggressive
profile-guided quantization fails at 1B scale) and treat cross-silicon
validation on Jetson, Apple Silicon, and consumer RTX as the natural
follow-up, enabling LLM deployment across previously infeasible memory
regimes without modifying model weights.
\end{abstract}

\section{Introduction}
\label{sec:intro}

The operational question in LLM inference today is not only how much a
model can be compressed but which devices can run it at all. A 4-bit 8B
model still requires $\sim$5--6\,GB of resident VRAM; a 3B model in BF16
requires $\sim$7\,GB. These footprints place most consumer GPUs, mobile
accelerators, integrated graphics, and older cloud silicon outside the
feasible region. Existing post-training quantization
commits the precision choice at calibration time and emits a new set of
weights keyed to that choice, so the set of viable targets for a given
model is fixed before deployment. In a setting where target hardware is
heterogeneous and individual device budgets vary across sessions
(thermal throttling, battery state, other processes), this binding is
the limiting factor.

We move the per-layer precision decision from calibration time to load
time. A short forward pass over a handful of calibration prompts is
enough to rank layers by their sensitivity to precision reduction; the
ranking is computed on the target device in seconds, cached by
architecture, and reused across every subsequent load. From that single
signal we drive two per-layer decisions: precision dispatch (W4A8
vs.\ W4A16) and residency tier (GPU / CPU RAM / SSD). The same weights
serve multiple memory budgets, and the set of reachable operating
points widens accordingly.

\paragraph{Contributions.}

\begin{enumerate}[leftmargin=*, itemsep=2pt]
\item \textbf{MCAP, a load-time per-layer importance signal}
  (Section~\ref{sec:mcap}). A 60-second, gradient-free, weight-free
  profiler over 12 calibration prompts (floor of 8 for stable top-$k$
  recovery, tunable count and domain, extensible to live traffic).
  Scores are hardware-independent and cached per architecture. Across
  eight architectures from GPT-2 (0.1B) through Qwen2-7.6B, MCAP recovers
  a strongly non-uniform layer-importance pattern that a static ``protect
  the last layer'' rule does not reproduce.

\item \textbf{Importance-guided execution coupling precision and
  residency} (Section~\ref{sec:w4a8_kernel}, Section~\ref{sec:paging}).
  At decode time each layer is routed on its MCAP score to a precision
  tier (W4A8 or W4A16) and a residency tier (GPU, CPU RAM, or SSD),
  driven by one signal rather than two independent systems. The W4A8
  path delivers 2.3--2.9$\times$ over W4A16 and 1.5--1.8$\times$ over
  llama.cpp Q4\_0 on T4, with the gap widening at larger scale
  (Table~\ref{tab:throughput}). The three-tier pager runs models
  3.5--4$\times$ larger than GPU memory (Llama-3.2-3B in 2\,GB,
  Llama-3.1-8B in 4\,GB) with no observable degradation on evaluated
  tasks relative to the unconstrained baseline, and
  raises throughput by 9--58\% over the unconstrained run in several
  configurations.

\item \textbf{An integrated Rust+CUDA implementation spanning 12+
  architectures} (Section~\ref{sec:system}). 22K lines of Rust and
  1,413 lines of hand-fused CUDA (RMSNorm+RoPE, fused QKV projections,
  flash decode attention, W4A8, W4A16) behind a unified
  \texttt{GenericBlockWeights} abstraction covering Llama, Qwen, Phi,
  Gemma, GPT-2, Falcon, and GPT-NeoX. MCAP profiles and per-layer
  dispatch work unchanged across all supported architectures.
\end{enumerate}

\needspace{6\baselineskip}
\paragraph{Why the three components are coupled.} MCAP provides the signal
that makes adaptive execution possible; routing and paging are what turn
that signal into throughput and feasibility. Routing alone reduces to a
static rule that mis-fires on architectures where the rule is wrong, and
paging alone is LRU without an importance prior. The strongest results
appear where the three reinforce each other, in the regime they were
designed for: memory budgets at which conventional inference is infeasible.

\paragraph{Feasibility first, throughput second.} The primary result is
about which operating points are reachable. NVE reaches points that
GPU-resident inference cannot at any precision: Llama-3.2-3B with full
BF16 weights ($\sim$7\,GB) runs in 2\,GB with no observable
degradation on the evaluated task suite, and Llama-3.1-8B
($\sim$16\,GB BF16) runs in 4\,GB while matching its constrained
baseline on the same suite. Four-bit quantization alone still
requires $\sim$5--6\,GB for an 8B model, so the 4\,GB point is a new
hardware class rather than an incremental improvement. Throughput is
the secondary result: at matched arithmetic precision the W4A8 routing
path delivers 1.5--1.8$\times$ over llama.cpp Q4\_0, and W4A16, W4A8,
and MCAP Mixed show no measurable difference on WikiText-2 and
HellaSwag at the evaluated sample sizes.

\begin{figure}[H]
  \centering
  \includegraphics[width=\linewidth]{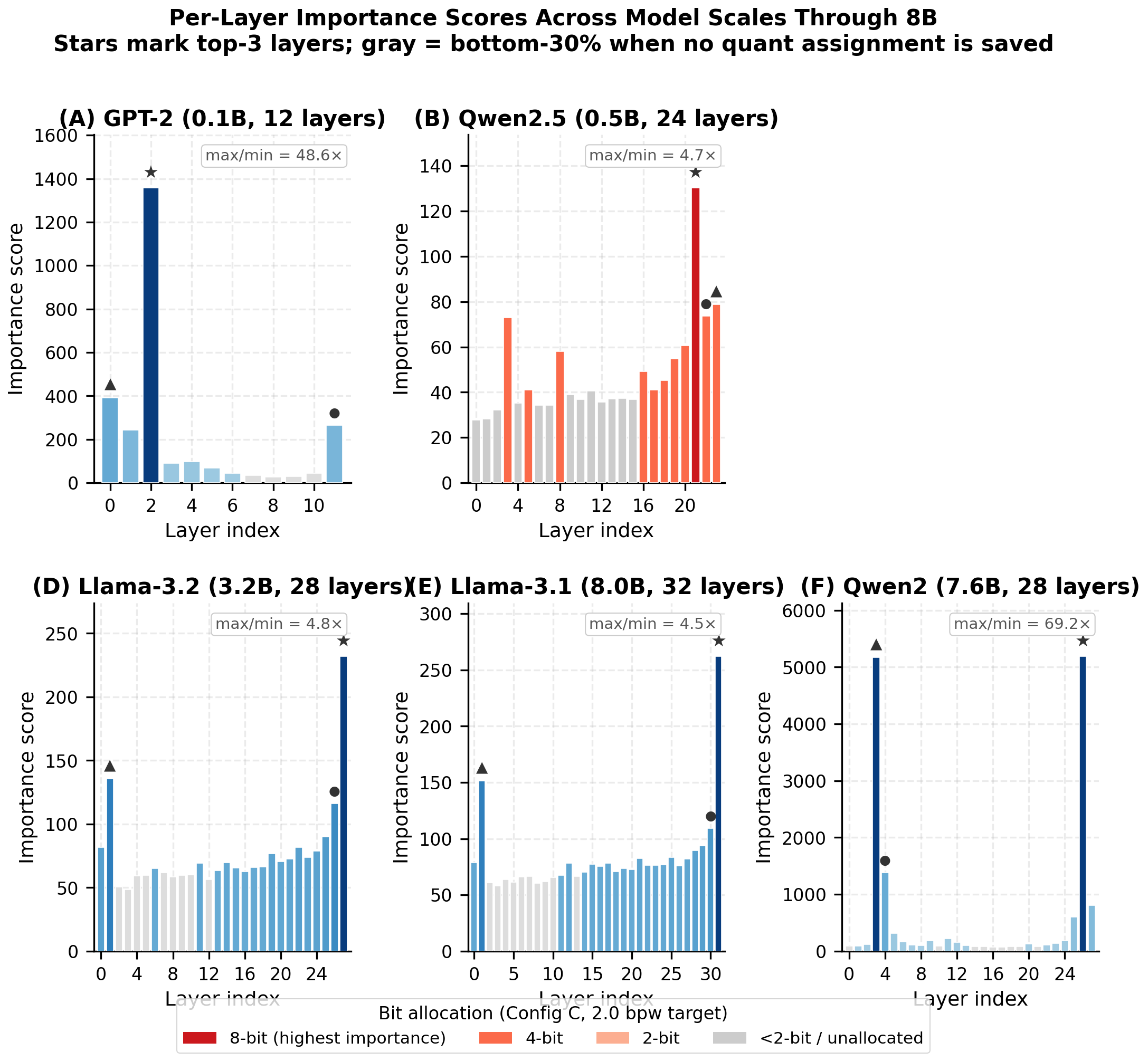}
  \caption{\textbf{MCAP importance scores across model scales through 8B.}
  Per-layer importance scores for GPT-2 (0.1B), Qwen2.5 (0.5B), Llama-3.2-1B,
  Llama-3.2-3B, Llama-3.1-8B, and Qwen2-7.6B. The dominant outlier layer is usually near
  the end of the network, but the exact pattern is architecture-dependent: most models show
  a single final-layer outlier, while larger Qwen variants exhibit additional outliers.}
  \label{fig:importance_4panel}
\end{figure}

Figure~\ref{fig:importance_4panel} makes the profiling claim concrete. The non-uniformity
is stable from 0.1B through 8B, but the \emph{pattern} is not rigid: most Llama-family
runs show one dominant late-layer peak; GPT-2 and larger Qwen variants show several.
This is the reason a runtime signal is preferable to a static rule such as
``always protect the last layer'': the rule is correct on average but wrong on
the architectures where it is wrong.

\paragraph{Quantitative consequences.} On NVIDIA T4, the throughput gap
corresponds to a 44\% reduction in tok/s$^{-1}$ cost vs.\ llama.cpp at 1B
(Section~\ref{sec:throughput}). The W4A8 advantage over W4A16 grows from
2.3$\times$ at 1B to 2.9$\times$ at 8B, consistent with a bandwidth-limited
regime rather than a small-model effect. The memory side reaches points not
accessible to GPU-resident inference. The same MCAP signal drives three
execution modes on a single spectrum (Section~\ref{sec:deployment_modes}):
\emph{hot-only} (skip low-importance layers entirely; no paging traffic),
\emph{hot-only+AWQ} (saliency-weighted quantization on the retained layers for
higher quality when the budget fits), and \emph{paging} (GPU$\to$RAM$\to$SSD)
for budgets below the hot-only viability floor, reaching full-BF16 3B in 2\,GB
and 8B in 4\,GB with no observable degradation on the evaluated task suite
(Section~\ref{sec:memory}).

\section{Background and Related Work}
\label{sec:related}

This paper focuses on the load-time decision problem: how per-layer
precision and memory placement should be chosen for a given set of
weights under varying hardware constraints. We survey two threads that
bear directly on that problem: post-training quantization (which decides
precision at calibration time) and weight offloading / memory-efficient
inference (which decides placement, usually at inference time). The
contribution of this paper is the coupling of these two decisions on a
single load-time signal.

\subsection{Post-Training Quantization for LLMs}

Post-training quantization (PTQ) compresses model weights, activations,
and increasingly the KV cache after training, to reduce memory and
compute. The field has advanced rapidly in 2022--2024 toward full-stack
quantization (weights + activations + KV cache) rather than weight-only
compression, and several methods already use runtime-aware components.
We summarize the lines of work most relevant to this paper.

\paragraph{Weight-only quantization.} GPTQ~\cite{frantar2022gptq} uses
approximate second-order information (Hessian inverse) to find per-column
quantization scales, achieving 3--4 bit precision with small perplexity
degradation; it is explicitly designed to scale to very large models
efficiently. AWQ~\cite{lin2023awq} observes that $\sim$1\% of weight channels are
disproportionately important and protects them via per-channel scaling;
AWQ already uses activation statistics to identify salient channels and
explicitly argues for cross-domain generalization without backprop or
reconstruction. Its saliency signal is a form of importance estimation
at finer granularity than ours. More recent methods use rotation-based outlier
suppression: QuaRot~\cite{ashkboos2024quarot} applies Hadamard rotations to
flatten outliers before 4-bit quantization, and
SpinQuant~\cite{liu2024spinquant} learns the rotation matrices end-to-end.
HQQ~\cite{badri2023hqq} uses half-quadratic splitting for calibration-free
weight quantization. Marlin~\cite{frantar2024marlin} is a reference
high-throughput W4A16 GEMM kernel on Ampere/Hopper and co-designs the
kernel with the weight format. These methods produce static quantized
weights but often couple to online components at inference time
(per-token activation handling, runtime scale application).

\paragraph{Activation quantization.} Activation behavior, rather than weights
alone, is the central bottleneck in much of the recent PTQ literature.
SmoothQuant~\cite{xiao2022smoothquant} argues that activations are harder to
quantize than weights and migrates the difficulty between the two via a
mathematically equivalent per-channel scaling that is applied online.
LLM.int8()~\cite{dettmers2022llm} performs mixed-precision decomposition
for outlier dimensions \emph{during inference}: outlier features
($>$6$\sigma$) use FP16 while the remaining $>$99.9\% use INT8. This is
runtime outlier handling at the feature-dimension level, not a runtime
profiling scheme at the layer level. Atom~\cite{zhao2024atom} co-designs
low-bit serving with mixed-precision reordering, dynamic activation
quantization, low-bit kernels, and KV-cache quantization for W4A4, and
is the closest published W4A8-adjacent kernel to ours. Atom is stronger
prior art on low-bit serving co-design; it does not perform runtime
layer-level importance profiling, which is the axis we target.
ZeroQuant-V2~\cite{yao2023zeroquantv2} reports that activation quantization
is more sensitive than weight quantization and studies layer-wise sensitivity
for mixed-precision assignment; OmniQuant~\cite{shao2024omniquant} learns
transformations that shift quantization difficulty between weights and
activations. KIVI~\cite{liu2024kivi} targets the KV cache, which can exceed the model
weight footprint in long-context settings; it is adjacent to our work
rather than directly competing with the weight-residency mechanism we use. Two features of this line of work are relevant here: activation magnitude
and outlier structure are central to modern PTQ, and calibration is
typically combined with online components (equivalent scaling at runtime,
dynamic per-token activation quantization, runtime outlier
decomposition). Our contribution is that a single load-time
per-\emph{layer} signal can drive both precision dispatch and memory
residency without per-target calibration.

\paragraph{Weight-sensitivity methods.} SqueezeLLM~\cite{kim2024squeezellm} computes
per-weight sensitivity from the Hessian and uses it for non-uniform quantization with a
dense-sparse decomposition; SpQR~\cite{dettmers2024spqr} isolates outlier
weights offline and stores them separately. Both operate at per-weight
granularity, so importance is already a well-studied quantity in PTQ; MCAP
is orthogonal, operating at the coarser per-layer granularity on unmodified
weights and driving paging as well as precision.

\paragraph{Per-layer bit allocation.} The EXL2
format~\cite{turboderp2024exl2} used by ExLlamaV2 already implements
per-layer bit allocation, chosen from offline error measurements; based on
its public documentation it is the closest existing work to the per-layer
granularity we use. MCAP differs in three ways: it computes the per-layer signal at load
time, it selects precision per activation path (W4A8 vs.\ W4A16) rather
than per weight, and the same scores drive paging placement. EXL2 chooses
bits per weight offline; MCAP chooses per-activation precision and
residency tier at load time.

\paragraph{KV cache and co-design.} Model weights are not the only
memory-dominant object in modern LLM inference. KIVI~\cite{liu2024kivi} and
related work show that the KV cache can exceed the model weight footprint in
long-context settings. Several prior systems already co-design kernels or
memory systems with quantization: MARLIN~\cite{frantar2024marlin} co-designs
a high-throughput W4A16 GEMM with quantization format, Atom co-designs
kernels and serving behavior for W4A4, and KIVI co-designs KV-cache
quantization with memory management. Our work focuses on per-layer
weight-precision and weight-residency coupling, and treats KV-cache
management as an orthogonal axis that can be composed with our signal
(we do not quantize the KV cache in this paper; a natural extension is
MCAP-guided KV precision, noted in the Conclusion).

\paragraph{Gap.} Prior work has studied importance at weight, channel,
and (in EXL2's case) layer granularity, as well as activation outliers at
runtime. What prior work does not provide is a load-time per-layer signal
that (a)~is computed on the target device from unmodified weights,
(b)~is cheap enough to recompute when conditions change, (c)~drives
precision dispatch and memory residency from a single decision, and
(d)~allows the same calibrated weights to serve different memory budgets.
MCAP introduces a \emph{load-time control layer} for LLM inference: a
single per-layer signal, computed on the target device, that couples the
precision and residency decisions existing systems make separately. This
layer sits above within-layer quantization methods (GPTQ, AWQ,
SmoothQuant, OmniQuant), per-weight bit allocation formats (EXL2), and
KV-cache quantization (KIVI), all of which remain usable underneath.

\subsection{LLM Inference Engines and Weight Offloading}

\textbf{llama.cpp}~\cite{ggerganov2023llamacpp} pioneered CPU-first inference with GGML
quantization and later GPU offload. \textbf{vLLM}~\cite{kwon2023efficient} introduced
PagedAttention for KV (key/value) cache management and supports GPTQ/AWQ/FP8 weights, but its
paged-attention abstraction is for KV cache, not weight virtualization.
\textbf{TensorRT-LLM}~\cite{nvidia2023tensorrt} provides NVIDIA-optimized static
compilation but requires offline model conversion. \textbf{ExLlamaV2}~\cite{turboderp2023exllama}
ships fast GPTQ/EXL2 inference kernels including mixed per-layer bit allocation (via the
EXL2 format~\cite{turboderp2024exl2}).

\paragraph{Weight offloading and paging.} FlexGen~\cite{sheng2023flexgen}
demonstrates high-throughput LLM inference on a single GPU by offloading
weights, activations, and KV cache across GPU/CPU/disk, optimizing batch
throughput. PowerInfer~\cite{song2024powerinfer} exploits activation
sparsity to keep ``hot'' neurons on GPU and ``cold'' neurons on CPU; its
neuron-level routing is finer-grained than the per-layer routing we use and
is the closest prior art to our importance-guided tiering.
LLM-in-a-Flash~\cite{alizadeh2024llmflash} streams weights from flash
storage on memory-constrained devices. These systems establish that
weight offloading is viable on its own; our work builds on them by coupling
layer-level importance profiling with on-GPU mixed-precision dispatch in a
single latency-oriented engine.

\paragraph{Positioning MCAP against prior work.} The nearest research neighbors are
not PTQ systems but edge-focused inference work: LLM-in-a-Flash~\cite{alizadeh2024llmflash}
streams weights from flash on memory-constrained devices, and
PowerInfer~\cite{song2024powerinfer} uses a learned sparsity predictor to route neurons
between GPU and CPU. Both target the deployment regime MCAP addresses, and both make
different tradeoffs (Table~\ref{tab:pipeline_properties}).

\paragraph{MCAP vs.\ PowerInfer, in detail.} PowerInfer is the closest
comparison on the memory-routing axis. Three concrete differences: (a)~\textbf{no
trained predictor}. PowerInfer trains a per-model MLP sparsity predictor that
runs alongside inference to decide which neurons to activate; the predictor is
an additional file that must be trained and kept in sync with the weights. MCAP
is a deterministic forward pass over 12 prompts that produces a 16--32 float
JSON cached per architecture. (b)~\textbf{Layer granularity, not neuron.}
PowerInfer's neuron-level routing is finer-grained but requires a predictor at
inference time; MCAP's layer-level decisions are made once at load and amortized
over all subsequent tokens. (c)~\textbf{One signal, both axes.} PowerInfer
routes weights for residency only; it does not address precision. MCAP drives
precision dispatch \emph{and} memory residency from the same score, so no
separate quantization strategy is needed. Concretely, a PowerInfer configuration
requires $\{$weights, trained predictor, quantized weights$\}$; an MCAP
configuration requires $\{$weights, per-architecture profile JSON$\}$ and
computes the per-device precision/residency plan at load. The predictor is also
the piece most sensitive to cross-device variation: trained on one distribution,
it must be retrained when the base weights change, whereas a deterministic
forward-pass profiler reflects that update automatically.

The axis we contribute is a load-time, architecture-keyed control signal
that drives precision and residency simultaneously, without a per-model
trained predictor. Within-layer quantization continues to run underneath:
Config~C uses AWQ-style scaling on the retained layers and composes with
MCAP routing directly.

\begin{table}[H]
\centering
\caption{\textbf{Configuration properties across memory-efficient LLM systems.}
MCAP is the only approach that both avoids weight modification and avoids a
per-model trained predictor, while driving precision and residency from a
single signal.}
\label{tab:pipeline_properties}
\footnotesize
\setlength{\tabcolsep}{3pt}
\begin{tabular}{lccccc}
\toprule
System & Offline & Configuration & Recal. & Per-model & Drives \\
       & cost    &                  & cost   & predictor & prec.+res.? \\
\midrule
GPTQ~\cite{frantar2022gptq}        & hours       & modified weights     & hours  & $\times$    & prec. \\
AWQ~\cite{lin2023awq}              & hours       & modified weights     & hours  & $\times$    & prec. \\
EXL2~\cite{turboderp2024exl2}      & hours       & modified weights     & hours  & $\times$    & prec. \\
PowerInfer~\cite{song2024powerinfer} & train+inf. & weights + MLP & retrain & \checkmark & res. \\
LLM-in-Flash~\cite{alizadeh2024llmflash} & ---  & original weights     & ---    & $\times$    & res. \\
FlexGen~\cite{sheng2023flexgen}    & ---         & original weights     & ---    & $\times$    & res. \\
\textbf{MCAP (ours)}               & \textbf{60\,s} & \textbf{orig.\ weights + JSON} & \textbf{60\,s} & $\times$ & \textbf{both} \\
\bottomrule
\end{tabular}

\smallskip
{\footnotesize \textbf{``Per-model predictor''} refers to an inference-time
neural network trained per base model on representative data and retrained when
the base model changes. PowerInfer's MLP sparsity predictor is the canonical
example. A forward-pass profiler like MCAP produces a deterministic
byte-identical JSON from the weights alone, so no predictor exists to drift.}
\end{table} The focus of this paper is the load-time decision signal; the
W4A8 dp4a kernel and the three-tier pager make that signal actionable.
Evaluation spans 0.1B--8B, with throughput on Llama 1B/3B/8B and
memory-constrained deployment through 8B; cross-silicon validation on
non-NVIDIA targets is follow-up work (Section~\ref{sec:limitations}).

\subsection{Concurrent and Orthogonal Directions (2024--2026)}
\label{sec:landscape}

The 2024--2026 literature is converging on a shared premise, ``keep
only what matters resident, move the rest,'' with each line of work
differing in \emph{how} that decision is made. Our contribution is the
load-time decision signal; the neighbors below focus on the mechanism
that acts on such a signal.

\textbf{Activation sparsity, neuron-level routing.}
PowerInfer~\cite{song2024powerinfer} and PowerInfer-2~\cite{xue2024powerinfer2}
are the closest conceptual neighbors: both exploit neuron-level activation
sparsity and route hot neurons to fast memory via a \emph{trained MLP
predictor} produced per base model. MCAP differs on three axes already expanded above:
layer-level (not neuron-level) granularity, no trained predictor, and a single
signal driving both precision and residency rather than residency alone.

\textbf{Flash- and SSD-resident weights.}
LLM-in-a-Flash~\cite{alizadeh2024llmflash} and FlexGen~\cite{sheng2023flexgen}
treat offloading as a data-movement problem, using access-pattern
heuristics and async prefetching. MCAP reframes the same problem as a
decision problem: the importance score determines what is worth moving
in the first place. The movement machinery from these systems can sit
underneath such a signal.

\textbf{KV-cache offloading.}
HeadInfer~\cite{luo2025headinfer} offloads KV cache per attention head, achieving
$\sim$92\% KV memory reduction for million-token contexts, and
InstInfer~\cite{pan2024instinfer} pushes KV cache into storage-class devices.
These target an orthogonal axis (KV cache) to MCAP (weights + execution strategy);
the two can be composed in a single deployment.

\textbf{Speculative decoding.}
Speculative decoding~\cite{leviathan2023speculative,chen2023speculative} accelerates
the \emph{token} loop via draft-and-verify and is now standard in vLLM and
TensorRT-LLM. MCAP operates at the \emph{model-execution} layer, below the token
loop, so the two stack: a speculatively decoded model still benefits from
importance-driven precision and residency on each verified step.

\textbf{Extreme low-bit training.}
BitNet~/1.58-bit LLMs~\cite{ma2024bitnet158} change the model \emph{representation}
at training time, achieving near-FP16 quality at $\sim$1--2 bits. This is a
training-time alternative; MCAP is a deployment-time control layer applied to
already-trained weights, and the two are not mutually exclusive, a BitNet model
can still be profiled by MCAP to steer which blocks pay additional precision cost.

\textbf{Adaptive runtime scheduling.}
ActiveFlow~\cite{zhang2025activeflow} dynamically swaps weights based on runtime
usage, and NEO~\cite{jiang2025neo} schedules CPU offloading via pipeline-balance
heuristics. Both are dynamic but heuristic- or architecture-driven; MCAP supplies
an explicit, per-layer importance signal that such schedulers could consume as
input rather than re-derive at runtime.

\paragraph{Summary.} Across these directions, the open question is
\emph{what decides what matters}: a learned predictor (PowerInfer), an
access heuristic (Flash/FlexGen), a structural partition (HeadInfer), a
draft model (speculative decoding), or training-time compression
(BitNet). MCAP's answer is a load-time, weight-free, architecture-keyed
importance signal that unifies sparsity, precision, and residency
decisions within a single control layer above these mechanisms.

\section{NVE System Design}
\label{sec:system}

\begin{figure}[H]
\centering
\resizebox{\linewidth}{!}{%
\begin{tikzpicture}[
    box/.style={draw, rounded corners=3pt, minimum width=4.5cm, minimum height=2.2cm,
                font=\small, align=center, thick},
    smallbox/.style={draw, rounded corners=2pt, minimum width=3.2cm, minimum height=0.6cm,
                     font=\scriptsize, align=center},
    arr/.style={-{Stealth[length=2.5mm]}, thick},
    label/.style={font=\scriptsize\itshape, text=nvegray},
]
\node[box, fill=nvepurple!15] (profiler) at (0, 0) {
  \textbf{MCAP Profiler}\\[4pt]
  12 calibration prompts\\
  $\to$ per-layer scores\\[2pt]
  $s_i = \text{mean}_{j,t} \|a_{i,t}\|$\\[2pt]
  \textit{60 sec, $O(1)$ memory}
};
\node[box, fill=tier1!40] (pager) at (7, 0) {
  \textbf{Virtual Weight}\\
  \textbf{Pager (3-Tier)}\\[4pt]
  Tier 0: GPU VRAM\\
  Tier 1: CPU RAM\\
  Tier 2: SSD/NVMe\\[2pt]
  \textit{LRU + prefetch}
};
\node[box, fill=w4a8color!15] (dispatch) at (14, 0) {
  \textbf{GPU Dispatch}\\
  \textbf{Layer}\\[4pt]
  $\hat{s}_i \geq \tau$: W4A16\\
  $\hat{s}_i < \tau$: W4A8\\[2pt]
  \textit{17 fused CUDA kernels}
};
\draw[arr, nvepurple] (profiler.east) -- (pager.west)
  node[midway, above, label] {importance scores};
\draw[arr, nveblue] (pager.east) -- (dispatch.west)
  node[midway, above, label] {GPU-resident weights};
\node[smallbox, fill=nvegray!10] at (0, -2.2) {\texttt{\~{}/.cache/nve/importance/}};
\node[smallbox, fill=nvegray!10] at (7, -2.2) {\texttt{/dev/shm} (mmap)};
\node[smallbox, fill=nvegray!10] at (14, -2.2) {GPU VRAM (allocated)};
\draw[dashed, nvegray] (profiler.south) -- (0, -1.9);
\draw[dashed, nvegray] (pager.south) -- (7, -1.9);
\draw[dashed, nvegray] (dispatch.south) -- (14, -1.9);
\end{tikzpicture}%
}
\caption{\textbf{NVE system architecture.} MCAP Profiler computes per-layer importance;
Virtual Weight Pager assigns layers to GPU/RAM/SSD tiers; GPU Dispatch routes each layer
to W4A8 or W4A16 kernels based on the MCAP threshold.}
\label{fig:architecture}
\end{figure}
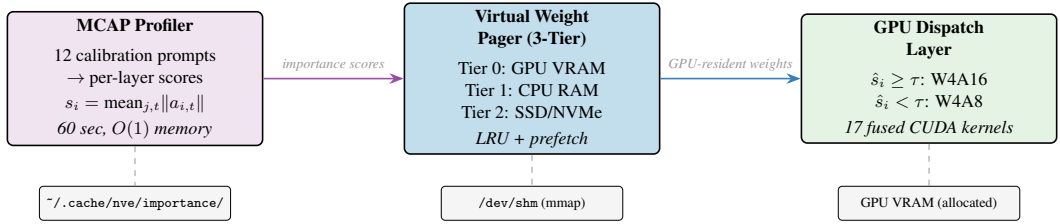

NVE realizes MCAP as a decision layer that drives three subsystems from one signal
(Figure~\ref{fig:architecture}): the MCAP Profiler computes per-layer importance, the
Virtual Weight Pager assigns layers to GPU/RAM/SSD tiers, and the GPU Dispatch Layer
routes each layer to W4A8 or W4A16 kernels. The engine is 22,000 lines of Rust with
1,413 lines of hand-tuned CUDA, and supports 12+ model architectures (Llama, Qwen, Phi,
Gemma, GPT-2, Falcon, GPT-NeoX) through a unified \texttt{GenericBlockWeights}
abstraction that handles fused QKV projections, Conv1D transposes, bias variants, and
fused gate+up projections transparently.

\subsection{MCAP: Monte Carlo Activation Profiler}
\label{sec:mcap}

MCAP produces a per-layer \emph{load-time control signal}, online and
without rewriting weights. It operates at coarser granularity than the
per-channel or per-weight sensitivity analyses in prior work (LLM.int8()
outlier detection, SqueezeLLM sensitivity scoring, AWQ activation-aware
scaling), which continue to run underneath it when applied.

\paragraph{Why ``Monte Carlo.''}
\begin{sloppypar}
MCAP estimates the per-layer activation expectation
$\mathbb{E}_{p \sim \mathcal{P}_{\text{text}}}\!\left[\frac{1}{|p|}\sum_t \|a_i(p, t)\|\right]$,
where $a_i(p,t) := \|[Q_i x_t, V_i x_t]\|_2 + \|\text{FFN}_i(x_t)\|_2$ is the combined
attn+FFN magnitude (Algorithm~\ref{alg:mcap}), via a sample mean over $k{=}12$
calibration prompts. This is a Monte Carlo estimator in the technical sense: a
sample-mean approximation of an expectation over an input distribution. The 12 prompts
are \emph{stratified} by topic (science, code, history, math) rather than drawn iid;
this is purposive calibration sampling, not a formal quasi-Monte Carlo construction
(no low-discrepancy sequence in a measurable sample space), and we use the terminology
informally. We verify convergence empirically through split-half overlap
(Section~\ref{sec:calibration}) and treat $k{=}12$ as an operationally chosen safety
margin rather than a theoretically derived one.
\end{sloppypar}

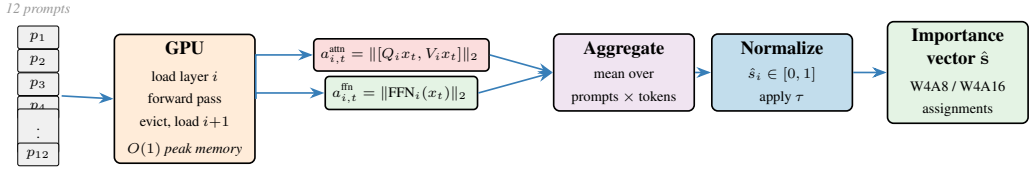
\begin{figure}[H]
\centering
\resizebox{0.98\linewidth}{!}{%
\begin{tikzpicture}[
    prompt/.style={draw, rounded corners=1pt, minimum width=0.7cm, minimum height=0.35cm,
                   font=\scriptsize, fill=nvegray!15},
    gpu/.style={draw, rounded corners=3pt, minimum width=2.4cm, minimum height=1.7cm,
                font=\small, align=center, fill=nveorange!15, thick},
    stat/.style={draw, rounded corners=2pt, minimum width=2.6cm, minimum height=0.55cm,
                 font=\scriptsize, align=center, thick},
    stage/.style={draw, rounded corners=3pt, minimum width=2.4cm, minimum height=1.1cm,
                  font=\small, align=center, thick},
    arr/.style={-{Stealth[length=2.5mm]}, thick, nveblue},
    note/.style={font=\scriptsize\itshape, text=nvegray},
]
\foreach \i in {1,...,4} {
  \node[prompt] at (1.3, 1.1-0.4*\i) {$p_\i$};
}
\node[prompt] at (1.3, 1.1-0.4*5) {$\vdots$};
\node[prompt] at (1.3, 1.1-0.4*6) {$p_{12}$};
\node[note, above=0pt] at (1.3, 0.95) {12 prompts};

\node[gpu] (gpu) at (3.8, -0.35) {
  \textbf{GPU}\\[2pt]
  \scriptsize load layer $i$\\
  \scriptsize forward pass\\
  \scriptsize evict, load $i{+}1$\\[2pt]
  \textit{\scriptsize $O(1)$ peak memory}
};

\node[stat, fill=nvered!15] (attn) at (7.5, 0.4) {
  $a_{i,t}^{\text{attn}} = \|[Q_i x_t, V_i x_t]\|_2$
};
\node[stat, fill=nvegreen!15] (ffn) at (7.5, -0.25) {
  $a_{i,t}^{\text{ffn}} = \|\text{FFN}_i(x_t)\|_2$
};

\node[stage, fill=nvepurple!15] (agg) at (11.3, 0.1) {
  \textbf{Aggregate}\\[1pt]
  \scriptsize mean over\\
  \scriptsize prompts $\times$ tokens
};

\node[stage, fill=tier1!40] (norm) at (14.0, 0.1) {
  \textbf{Normalize}\\[1pt]
  \scriptsize $\hat{s}_i \in [0,1]$\\
  \scriptsize apply $\tau$
};

\node[stage, minimum width=2.4cm, minimum height=1.4cm, fill=w4a8color!15] (out) at (17.0, 0.1) {
  \textbf{Importance}\\
  \textbf{vector $\hat{\mathbf{s}}$}\\[2pt]
  \scriptsize W4A8 / W4A16\\
  \scriptsize assignments
};

\draw[arr] (1.7, -0.3) -- (gpu.west);
\draw[arr] (gpu.east) |- (attn.west);
\draw[arr] (gpu.east) |- (ffn.west);
\draw[arr] (attn.east) -- (agg.west);
\draw[arr] (ffn.east)  -- (agg.west);
\draw[arr] (agg.east)  -- (norm.west);
\draw[arr] (norm.east) -- (out.west);

\end{tikzpicture}%
}
\caption{\textbf{MCAP streaming profiler dataflow.} Twelve calibration prompts pass
through the model one layer at a time: each layer is loaded to GPU, forwarded, and
evicted before the next is loaded, giving $O(1)$ peak memory with respect to depth
(Figure~\ref{fig:profiling_overhead}). Per-token attention and FFN magnitudes are
averaged across prompts, min-max normalized, and thresholded at $\tau$ to produce the
per-layer precision assignment consumed by all downstream components.}
\label{fig:mcap_dataflow}
\end{figure}

Concretely, MCAP assigns an importance score $s_i$ to each transformer layer
$i \in \{1, \ldots, L\}$ from the activation magnitudes above, streaming one layer at a
time through the forward pass so peak memory stays constant in depth
(Figure~\ref{fig:mcap_dataflow}). The key
intuition: INT8 quantization error is absolute, so layers whose activations occupy the
largest dynamic range are the ones that cannot absorb it.

\begin{algorithm}[t]
\caption{MCAP: Monte Carlo Activation Profiling}
\label{alg:mcap}
\begin{algorithmic}[1]
\REQUIRE Model $M$ with $L$ layers, calibration prompts $\mathcal{P} = \{p_1, \ldots, p_k\}$, threshold $\tau$
\ENSURE Per-layer importance scores $\mathbf{s} = (s_1, \ldots, s_L)$, precision assignments
\FOR{each prompt $p_j \in \mathcal{P}$}
  \STATE Run forward pass through $M$ on $p_j$
  \FOR{each layer $i \in \{1, \ldots, L\}$, each token $t$}
    \STATE $a_{i,t}^{\text{attn}} \leftarrow \| [Q_i \cdot x_t,\; V_i \cdot x_t] \|_2$
    \COMMENT{Attention proxy}
    \STATE $a_{i,t}^{\text{ffn}} \leftarrow \| \text{FFN}_i(x_t) \|_2$
    \COMMENT{FFN magnitude}
  \ENDFOR
\ENDFOR
\STATE $s_i \leftarrow \frac{1}{k} \sum_{j=1}^{k} \frac{1}{|p_j|} \sum_{t=1}^{|p_j|}
  \left( a_{i,t}^{\text{attn}} + a_{i,t}^{\text{ffn}} \right) \quad \forall\, i$
\IF{$\max(\mathbf{s}) - \min(\mathbf{s}) < \epsilon$}
  \STATE $\hat{s}_i \leftarrow 0 \quad \forall\, i$
  \COMMENT{Degenerate uniform case: no outlier, route all layers to W4A8}
\ELSE
  \STATE $\hat{s}_i \leftarrow (s_i - \min(\mathbf{s})) \,/\, (\max(\mathbf{s}) - \min(\mathbf{s}))$
  \COMMENT{Min-max normalize to $[0,1]$}
\ENDIF
\STATE \textbf{Assign:} layer $i$ uses W4A16 if $\hat{s}_i \geq \tau$, else W4A8
\end{algorithmic}
\end{algorithm}

For each calibration prompt and token position, MCAP computes two statistics at every layer:
(1)~\textbf{Attention proxy} $a_{i,t}^{\text{attn}} = \|[Q_i \cdot x_t,\; V_i \cdot x_t]\|_2$:
the L2 norm of concatenated query and value projections, capturing attention sensitivity
without the $O(n^2)$ attention computation; and
(2)~\textbf{FFN magnitude} $a_{i,t}^{\text{ffn}} = \|\text{FFN}_i(x_t)\|_2$:
the L2 norm of the feed-forward output, capturing the magnitude of the residual update.

Scores are averaged across all tokens and prompts, then min-max normalized to $[0, 1]$.
A threshold $\tau = 0.7$ partitions layers into high-importance (W4A16) and low-importance
(W4A8). In practice, this usually identifies a single dominant late-network outlier, but
the full 0.1B--8B sweep also exposes informative exceptions, including GPT-2's two-layer
pattern and Qwen2-7.6B's multi-outlier regime (Table~\ref{tab:mcap_scores},
Figure~\ref{fig:importance_4panel}).

\begin{table}[H]
\centering
\caption{\textbf{MCAP importance score distribution across 8 model scales.} Layer
importance is highly non-uniform throughout the 0.1B--8B sweep. Most models exhibit a
single dominant late-layer outlier, but GPT-2 and Qwen2-7.6B show multi-outlier patterns.
Split-half top-$k$ overlap is 1.0 for every model, which is consistent with stable
rankings but is a single point estimate on $n{=}12$ prompts with no confidence interval;
the large outlier gaps ($1.6\times$ to $39\times$) make top-1 recovery nearly forced, so
this test is strong evidence for stability of the leading outliers and weaker evidence
for mid-rank ordering. ``Ratio'' is the outlier score divided by the mean of the
remaining layers (i.e.\ the gap that top-1 recovery has to resolve), not max/min of
``Score Range.''}
\label{tab:mcap_scores}
\small
\begin{tabular}{lccccc}
\toprule
Model & Layers & Score Range & Outlier & Ratio & W4A16/W4A8 \\
\midrule
GPT-2 (0.1B) & 12 & 35--1301 & L3$^*$ (4.9$\times$) & 4.9$\times$ & 2 / 10 \\
Qwen2.5 (0.5B) & 24 & 28--130 & L22 (2.1$\times$) & 2.1$\times$ & 1 / 23 \\
Llama-3.2 (1.2B) & 16 & 61--142 & L16 (1.9$\times$) & 1.9$\times$ & 1 / 15 \\
Qwen2 (1.5B) & 28 & 103--224 & L28 (1.6$\times$) & 1.6$\times$ & 1 / 27 \\
Llama-3.2 (3.2B) & 28 & 48--232 & L28 (2.8$\times$) & 2.8$\times$ & 1 / 27 \\
Qwen2 (3.1B) & 36 & 52--936 & L36 (6.4$\times$) & 6.4$\times$ & 1 / 35 \\
Llama-3.1 (8B) & 32 & 58--262 & L32 (2.8$\times$) & 2.8$\times$ & 1 / 31 \\
Qwen2 (7.6B) & 28 & 75--5193 & L27 (39$\times$) & 39$\times$ & 3 / 25 \\
\bottomrule
\end{tabular}
\vspace{2pt}
{\footnotesize $^*$GPT-2 has two outliers (L1 and L3) due to its unique Conv1D + fused QKV
architecture. All other models have a single dominant outlier at the final layer.}
\end{table}

\paragraph{Why not hardcode ``last layer $=$ W4A16''?} The final layer is
consistently the outlier across Llama and Qwen families, but the pattern is not
universal: GPT-2 has two outliers at layers 1 and 3, Qwen2-7.6B has outliers at
layers 4, 5, and 27, and the outlier ratio varies from 1.6$\times$ to
39$\times$ across models. A runtime signal identifies the correct precision
assignment for each architecture without manual per-model analysis, at a cost
of roughly 60 seconds. Figure~\ref{fig:importance_4panel} illustrates the
pattern: what generalizes across scale is non-uniformity, not a single
universal layer index.

\begin{table}[H]
\centering
\caption{\textbf{MCAP normalized scores for Llama-3.2-1B.} All 16 layers, min-max
normalized to $[0, 1]$ (from the threshold ablation study, Appendix~\ref{app:reproducibility}). Only L15 exceeds the
$\tau = 0.7$ threshold for W4A16 dispatch. Note L1 (0.699) sits just below the threshold
and L14 (0.475) is the third-highest: even for Llama-family models with a dominant
final-layer outlier, the mid-network distribution is not flat. Hardcoding ``protect only
the last layer'' would ignore the gradient of importance across the rest of the network.}
\label{tab:per_layer_scores}
\footnotesize
\setlength{\tabcolsep}{4pt}
\begin{tabular}{lcccccccccccccccc}
\toprule
Layer & L0 & L1 & L2 & L3 & L4 & L5 & L6 & L7 & L8 & L9 & L10 & L11 & L12 & L13 & L14 & L15 \\
Score & 0.30 & 0.70 & 0.00 & 0.07 & 0.12 & 0.08 & 0.09 & 0.03 & 0.07 & 0.27 & 0.19 & 0.24 & 0.32 & 0.38 & 0.48 & \textbf{1.00} \\
\bottomrule
\end{tabular}
\end{table}

\paragraph{MCAP vs.\ GPTQ/AWQ calibration.} MCAP uses 12 diverse calibration
prompts by default; GPTQ typically uses 128. The two are measuring different
quantities: GPTQ estimates per-weight quantization sensitivity, which is a
high-dimensional target that benefits from more samples, while MCAP
estimates an $L$-dimensional per-layer signal, which concentrates faster.
Ablation shows score stability (top-$k$ overlap = 1.0) with as few as 8
prompts, so 12 is a safety margin rather than a tuned hyperparameter.
Profiles are cached and hardware-independent: a profile computed on one GPU
loads unchanged on another. This does not make MCAP a replacement for
GPTQ/AWQ calibration, which produces substantially lower PPL at matched
bit-widths; the two signals target different granularities.

The prompt set is a tunable parameter rather than a fixed design choice. As in
GPTQ/AWQ, where calibration sample count and domain materially affect
quantization quality, MCAP exposes the same controls, but at runtime rather
than embedded in weights: (a)~\textbf{count} can be increased when additional
statistical stability is required or reduced on latency-bound targets (the
8-prompt stability floor permits profiling in under 40\,s); (b)~\textbf{prompt
distribution} can be made domain-specific (code, math, multilingual, medical)
to bias importance toward the target workload, so a code-assistant use of the
same weights can carry a different MCAP profile from a general-chat use;
(c)~\textbf{task alignment} can use recent request traffic as the profiling
set, producing a profile that tracks the workload rather than a generic
corpus. This is the same calibration-design lever that existing PTQ methods
rely on, moved from a one-time offline step to a runtime-recomputable signal.

\paragraph{Amortization of the profiling cost.} The 60-second profiling cost is
paid once per architecture, not per run. MCAP profiles are cached at
\texttt{\textasciitilde/.cache/nve/importance/} keyed by model architecture
hash, so a profile computed once is reused by every subsequent load of the
same model, regardless of target hardware, and the cache invalidates only when
the underlying weights change. The amortized cost per inference run is
negligible. By contrast, GPTQ/AWQ rewrite weights and therefore produce a
separate set of quantized weights per precision target, which must be stored
and distributed alongside the original model.

\paragraph{Activation magnitude as a heuristic proxy.} Activation magnitude
is a heuristic for precision sensitivity rather than a principled guarantee:
since INT8 quantization error is absolute, layers whose query/value or FFN
outputs occupy larger dynamic ranges are more likely to incur meaningful
distortion when mapped to a fixed low-precision grid. AWQ uses a similar
empirical observation at the channel level. MCAP does not estimate the exact
downstream loss change from quantizing each tensor; it produces a
per-layer ordering that empirically correlates with precision sensitivity in
the tested architectures. MCAP is a lightweight runtime estimator of per-layer precision
sensitivity; it operates at a different granularity from a PTQ
optimizer, and its defining property is that it is recomputable at
load time.

\begin{figure}[H]
  \centering
  \includegraphics[width=\linewidth]{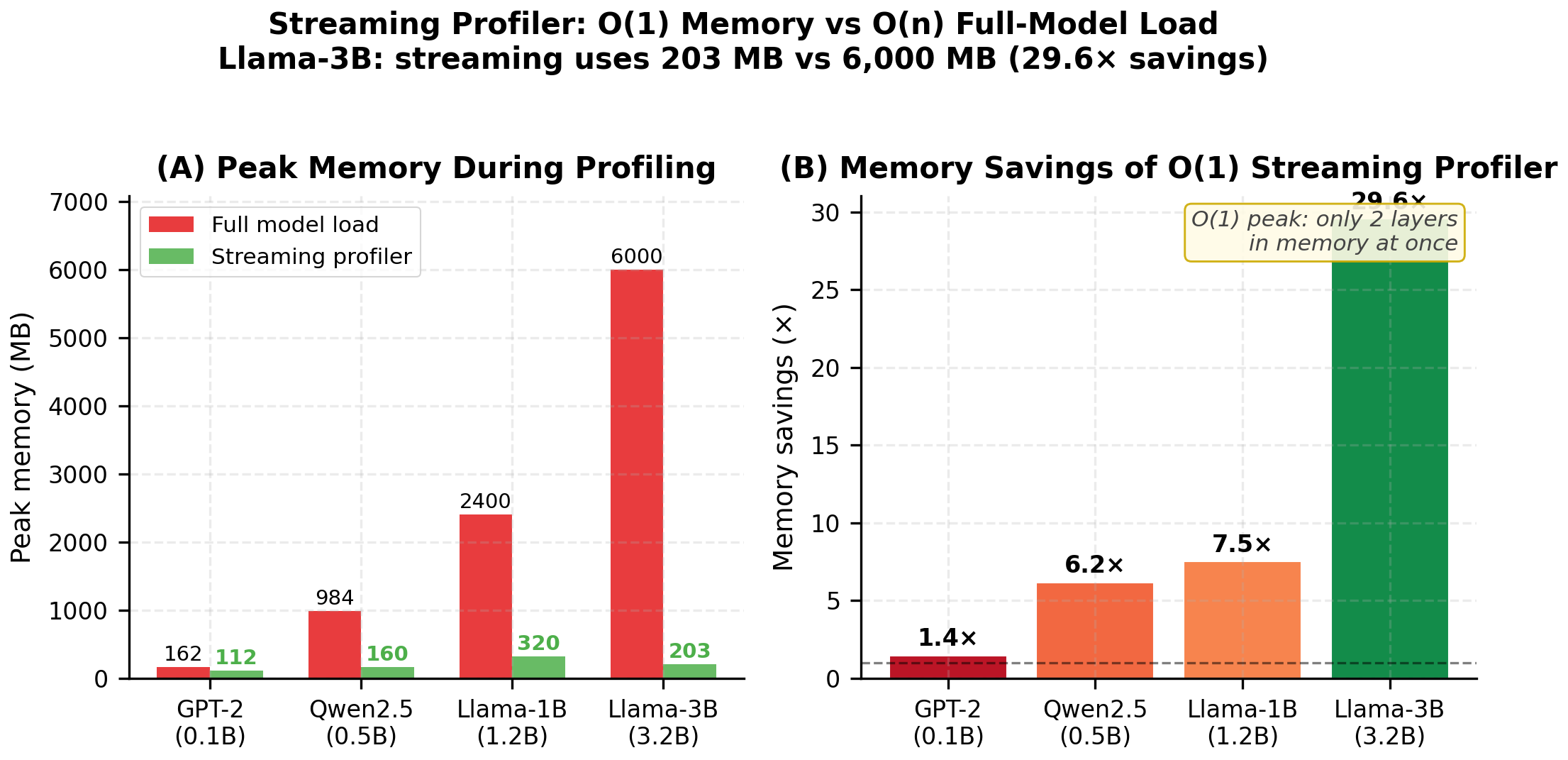}
  \caption{\textbf{MCAP streaming profiler memory advantage.} Peak memory comparison
  between full-model loading and MCAP's streaming approach. At 3B, the streaming profiler
  uses 29.6$\times$ less peak memory (203\,MB vs.\ 6,000\,MB), enabling profiling on
  devices that cannot load the full model.}
  \label{fig:profiling_overhead}
\end{figure}

\subsubsection{Recovery Guarantee: Scaling of Prompt Count with Outlier Gap}
\label{sec:mcap_theory}

The choice of 12 prompts is motivated by a concentration bound rather than
tuned empirically. MCAP is a \emph{selection} problem rather than a
rate-distortion problem, so the appropriate guarantee is top-$k$ recovery
rather than a Shannon-style distortion bound. This is orthogonal to, and
compatible with, recent work on information-theoretically optimal
per-coordinate quantization such as TurboQuant~\cite{zandieh2025turboquant},
which addresses \emph{how} to quantize each layer; MCAP addresses \emph{which}
layers to protect.

\paragraph{Assumption (sub-Gaussian activation norms).} For each layer $i$, let $s_i$
denote the population-level MCAP score (expected attention-plus-FFN L2 norm over the
prompt distribution), and let $\hat{s}_i^{(N)}$ denote the empirical estimate from $N$
calibration prompts. We assume that per-prompt activation norms are sub-Gaussian with
parameter $\sigma_a$ around their mean, i.e.,
$\Pr[|\hat{s}_i^{(N)} - s_i| \geq t] \leq 2\exp(-Nt^2 / 2\sigma_a^2)$. This assumption is
empirically justified by the per-layer norm variance we measure across the 12 calibration
prompts, which is tightly bounded for all 8 models from GPT-2 to Llama-3.1-8B.

\paragraph{Proposition (top-$k$ recovery).} Let $\Delta_k := s_{(k)} - s_{(k+1)}$ be the
gap between the $k$-th and $(k{+}1)$-th population-level scores in sorted order. Under the
above assumption, the empirical ranking $\hat{s}^{(N)}$ recovers the true top-$k$ set with
probability at least
\[
  1 - 2\,k(L-k) \cdot \exp\!\left(-\frac{N \Delta_k^2}{8 \sigma_a^2}\right),
\]
where $L$ is the number of layers.

\paragraph{Proof sketch.} A ranking inversion between any two layers $i, j$ with
$s_i > s_j$ requires $|\hat{s}_i^{(N)} - s_i| + |\hat{s}_j^{(N)} - s_j| \geq s_i - s_j$.
The probability of either deviation exceeding $(s_i - s_j)/2$ is bounded by the
sub-Gaussian tail. Union-bounding over the $k(L-k)$ pairs that straddle the top-$k$
boundary (whose minimum gap is $\Delta_k$) gives the stated bound. A full proof is
deferred to Appendix~\ref{app:mcap_proof}.

\paragraph{Corollary (scaling behavior).} Plugging in the measured values for
Llama-3.1-8B ($L=32$, $\Delta_1 / \sigma_a \approx 1.8$ from Table~\ref{tab:mcap_scores})
and $N=12$, the conservative union bound in Appendix~\ref{app:mcap_proof} yields a
a non-trivial but non-vacuous upper bound on the failure probability. The
qualitative content of the bound is that $N$ scales as
$\sigma_a^2 / \Delta_k^2$, so architectures with larger outlier gaps
(Qwen2-7.6B: $39\times$ ratio) require substantially fewer prompts than
architectures with tight gaps. The empirical top-$k$ recovery provides an
independent check: split-half top-$k$ overlap is 1.0 across all 8 tested
models (Table~\ref{tab:mcap_scores}), consistent with a tighter true
concentration than the worst-case Hoeffding bound certifies. We use the bound
as a scaling argument and the 1.0 split-half overlap as the empirical evidence
that 12 prompts are sufficient in practice.

\paragraph{Scope of the guarantee.} This proposition says that MCAP's \emph{ranking}
concentrates quickly; it does not say that the ranking is the globally optimal per-layer
precision assignment for minimizing end-task loss, that would require a distributional
link between activation magnitude and downstream perplexity under quantization, which no
published work establishes. The empirical evidence that MCAP Mixed matches W4A16 PPL in the tested
regimes (Tables~\ref{tab:wikitext}--\ref{tab:hellaswag}) is consistent with
the ranking being operationally correct in those regimes, but we state this as
evidence rather than a theorem. The recovery bound is what the current data
supports, and we do not claim beyond it.

\subsection{Per-Layer Precision Dispatch and the W4A8 Path}
\label{sec:w4a8_kernel}

The contribution in this subsection is the \emph{routing rule}: at each decode
step, layer $i$ is dispatched to W4A8 or W4A16 based on its normalized MCAP
score,
\[
  \text{route}(i) \;=\; \begin{cases} \text{W4A8} & \hat{s}_i < \tau \\ \text{W4A16} & \hat{s}_i \geq \tau \end{cases}
\]
The rule reduces to a single branch in the per-layer decode path and is
re-evaluated per layer, per forward pass. With $\tau{=}0.7$ on Llama-3.2-1B,
this routes one layer (Layer 16, the activation outlier) to W4A16 and the
remaining 15 to W4A8. The mechanism is straightforward; what makes it
effective is that MCAP provides an $\hat{s}_i$ that is both cheap to compute
(60\,s, no gradients, no weight modification) and accurate enough that the
majority of layers tolerate INT8 activations. The WikiText-2 and HellaSwag
equivalence results in Section~\ref{sec:quality} and the threshold-insensitivity
ablation in Section~\ref{sec:threshold} support this.

\begin{figure}[H]
\centering
\resizebox{0.85\linewidth}{!}{%
\begin{tikzpicture}[
    input/.style={draw, rounded corners=2pt, minimum width=2.4cm, minimum height=0.9cm,
                  font=\small, align=center, fill=nvegray!10, thick},
    decide/.style={draw, diamond, aspect=2.0, inner sep=2pt,
                   font=\small, align=center, fill=nvepurple!15, thick},
    path/.style={draw, rounded corners=2pt, minimum width=4.2cm, minimum height=0.9cm,
                 font=\small, align=center, thick},
    kernel/.style={draw, rounded corners=1pt, minimum width=4.0cm, minimum height=0.55cm,
                   font=\scriptsize\ttfamily, inner sep=2pt},
    arr/.style={-{Stealth[length=2.5mm]}, thick},
    edgelbl/.style={font=\scriptsize\itshape, text=nvegray},
]
\node[input] (layer) at (0, 0.6) {Layer $i$ input\\ \scriptsize (F16 activations $x_t$)};
\node[input, fill=nvepurple!10] (score) at (0, -0.8) {MCAP score $\hat{s}_i$};

\node[decide] (dec) at (3.6, -0.1) {$\hat{s}_i \geq \tau$?};

\node[path, fill=w4a16color!20] (p16) at (8.2, 2.2) {\textbf{W4A16 path}\\\scriptsize FP16 activations};
\node[kernel, fill=w4a16color!10] at (8.2, 1.4) {nve\_qkv\_matvec\_w4a16};
\node[kernel, fill=w4a16color!10] at (8.2, 0.8) {nve\_matvec\_w4a16};
\node[kernel, fill=w4a16color!10] at (8.2, 0.2) {nve\_flash\_decode\_f16};

\node[path, fill=w4a8color!20] (p8) at (8.2, -1.2) {\textbf{W4A8 path}\\\scriptsize INT8 activations (dp4a)};
\node[kernel, fill=w4a8color!10] at (8.2, -2.0) {nve\_quantize\_f16\_q8};
\node[kernel, fill=w4a8color!10] at (8.2, -2.6) {nve\_matvec\_w4a8};
\node[kernel, fill=w4a8color!10] at (8.2, -3.2) {deferred bias correction};

\node[input, minimum width=2.8cm] (tail) at (11.8, -0.5)
  {Layer $i{+}1$\\ \scriptsize (17 fused kernels total)};

\draw[arr, nveblue] (layer.east) -- ($(dec.west)+(0,0.3)$);
\draw[arr, nvepurple] (score.east) -- ($(dec.west)+(0,-0.3)$);
\draw[arr, w4a16color, thick] (dec.north east) -- (p16.west)
  node[edgelbl, midway, above, sloped] {yes ($\sim$1 layer/model)};
\draw[arr, w4a8color, thick] (dec.south east) -- (p8.west)
  node[edgelbl, midway, below, sloped] {no (typically $>$85\% of layers)};
\draw[arr, nvegray] (p16.east) -- ++(0.4, 0) |- (tail.west);
\draw[arr, nvegray] (p8.east)  -- ++(0.4, 0) |- (tail.west);

\node[font=\scriptsize\itshape, text=nvegray, anchor=north] at (3.6, -0.9)
  {$\tau = 0.7$ (normalized)};
\end{tikzpicture}%
}
\caption{\textbf{Per-layer precision dispatch.} At each decode step layer $i$ routes on
its normalized MCAP score: late-network activation outliers keep the FP16 path for
quality-critical quantities while the remainder (typically $>$85\% of layers, $15/16$
on Llama-3.2-1B) take the W4A8 dp4a path for $\sim$2$\times$ bandwidth. Every branch is a fused kernel
from the 17-kernel suite in Table~\ref{tab:kernels}.}
\label{fig:precision_dispatch}
\end{figure}
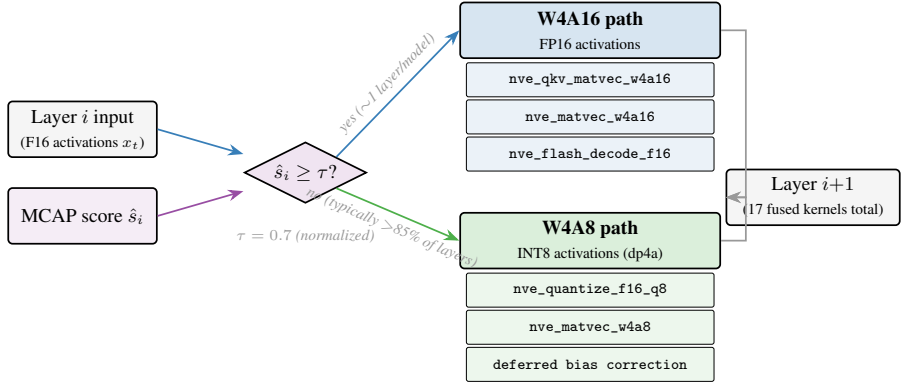

Figure~\ref{fig:precision_dispatch} shows the routing structure end-to-end: the
normalized MCAP score gates each layer into one of two kernel paths, both drawn from
the same fused suite.

The speedup obtained from this routing depends on the W4A8 path being fast.
INT8 dot-product kernels are not new: llama.cpp uses \texttt{\_\_dp4a}, and
Marlin and Atom target the same instruction family. NVE's decode path is
tuned for single-sequence decode and integrated into a graph-captured
pipeline, which accounts for the 1.5--1.8$\times$ gap over llama.cpp at
matched arithmetic precision (Section~\ref{sec:throughput}). The kernel
details below are engineering; the methodological point is that per-layer
routing on a runtime signal is what allows the INT8 path to be used without
measurable quality cost on non-outlier layers.

The W4A8 dp4a kernel drives Turing/Ampere INT8 dot-product instructions to
compute 4-bit weight $\times$ 8-bit activation matrix-vector products at roughly twice
the effective memory bandwidth of the W4A16 path.

\paragraph{Kernel architecture.} The \texttt{nve\_matvec\_w4a8} kernel computes one output
row per thread block with 4 warps (128 threads):

\begin{itemize}[leftmargin=*, itemsep=1pt]
\item \textbf{Activation quantization}: Input activations (F16) are dynamically quantized
  to INT8 per 32-element group via \texttt{nve\_quantize\_f16\_q8}, reused across Q/K/V
  projections within each layer.
\item \textbf{Weight format}: INT4 weights use the Q4\_0 block format (18 bytes per 32
  elements). Nibble layout follows the llama.cpp split format.
\item \textbf{dp4a accumulation}: Each \texttt{\_\_dp4a} performs
  $\text{acc} \mathrel{+}= \sum_{j=0}^{3} w_j \cdot x_j$ in INT32 (4 INT8 MACs per
  instruction); four such instructions process one Q4\_0 block of 32 weights.
\item \textbf{Deferred bias correction}: after accumulation,
  $\text{result} = s_w \cdot s_x \cdot (\text{sumi} - 8 \cdot \text{sum\_x})$,
  where $\text{sumi} = \sum_j w_j^{\text{uns}} \cdot x_j$ uses the unsigned nibble
  $w^{\text{uns}} \in [0, 15]$ and $\text{sum\_x} = \sum_j x_j$ carries the constant
  zero-offset correction (Q4\_0 centers the signed range at 8). This avoids a
  per-nibble subtract in the inner loop.
\item \textbf{Warp reduction}: 5-step \texttt{\_\_shfl\_xor\_sync} to scalar.
\end{itemize}

\paragraph{CUDA graph capture for decode.} The decode path is captured into a CUDA graph
and replayed per token, eliminating per-kernel launch overhead that otherwise accumulates
across 17 kernel invocations per layer. For a 16-layer 1B model at batch=1, this removes
roughly 0.6\,ms/token of launch overhead, material when the target per-token budget is
3--4\,ms. llama.cpp's default path issues kernels individually; this is a structural
reason the W4A8 throughput gap is not purely a kernel-efficiency effect. Graph capture
also fuses the KV-cache append pattern, eliminating 32 per-step allocations on 16-layer
models. The throughput numbers reported in Table~\ref{tab:throughput} include graph
capture in the NVE path; Table~\ref{tab:kernel_evolution}'s intermediate rows do not,
which is why the W4A8 row shows a step-function improvement rather than an incremental one.

\paragraph{Complete kernel suite.} NVE includes 17 custom fused CUDA kernels, each
replacing multiple unfused operations. Table~\ref{tab:kernels} lists the nine that are
load-bearing for decode throughput; the remaining eight are support kernels
(prefill-mode variants, embedding lookup, logit computation, residual add, token
sampling, scratch-buffer management) omitted here for space.

\begin{table}[H]
\centering
\caption{\textbf{NVE decode-path CUDA kernels.} Nine load-bearing fused kernels (of
17 total in the suite) that dominate decode throughput. All hand-tuned for NVIDIA T4
(sm\_75) with F16 I/O.}
\label{tab:kernels}
\small
\begin{tabular}{lll}
\toprule
Kernel & Replaces & Key Design \\
\midrule
\texttt{nve\_matvec\_w4a8} & cuBLAS F16 GEMV & dp4a INT8, 4 warps/row \\
\texttt{nve\_quantize\_f16\_q8} & --- (new) & Per-group absmax, 32-wide \\
\texttt{nve\_matvec\_w4a16} & cuBLAS F16 GEMV & Warp-shuffle, no dp4a \\
\texttt{nve\_qkv\_matvec\_w4a16} & 3$\times$ cuBLAS & Single launch, 3 projections \\
\texttt{nve\_flash\_decode\_f16} & 2 copies + 2 GEMM & GQA-native,$^{\ast}$ online softmax \\
\texttt{nve\_rms\_norm\_f16} & 4 ops & Single-pass, shared mem \\
\texttt{nve\_rope\_f16\_decode} & 3 ops & In-place, 1 block/head \\
\texttt{nve\_silu\_mul\_f16} & 2 ops & In-place on gate buffer \\
\texttt{nve\_dequant\_w4a16} & --- (prefill) & Materializes F16 for cuBLAS \\
\bottomrule
\end{tabular}

\smallskip
{\footnotesize $^{\ast}$GQA = grouped-query attention. GEMV/GEMM = general matrix-vector /
matrix-matrix multiply.}
\end{table}

\subsection{3-Tier Virtual Weight Paging}
\label{sec:paging}

The contribution here is architectural: NVE treats weight residency as a runtime
scheduling problem driven by learned layer importance, not as static placement or a
passive offload cache.

NVE implements a virtual memory system for transformer weights:

\begin{itemize}[leftmargin=*, itemsep=1pt]
\item \textbf{Tier 0 (GPU VRAM, ``hot'')}: Highest-importance layers. Zero-latency decode.
\item \textbf{Tier 1 (CPU RAM, ``warm'')}: LRU-cached. $\sim$1\,ms PCIe transfer.
\item \textbf{Tier 2 (SSD, ``cold'')}: On-demand. $\sim$10\,ms mmap load.
\end{itemize}

MCAP importance scores drive initial tier placement. Runtime LRU eviction and
importance-guided promotion maintain optimal residency. All models achieve $>$99.6\%
cache hit rate (Table~\ref{tab:paging_stats}).

\paragraph{Co-activation clustering via PMI.} Naive LRU treats each transformer
block as an independent paging unit, which is suboptimal: attention and FFN
sub-blocks within a single layer are co-activated on every token, and adjacent
layers are co-activated with high probability during autoregressive decode.
NVE computes pointwise mutual information between sub-block activations during
the MCAP calibration run and groups sub-blocks whose PMI exceeds a threshold
into atomic paging units. Promotions then operate at the cluster level, so a
demand-fault on one sub-block prefetches its co-activated neighbors in a
single PCIe transfer. Without co-activation prefetching, the observed miss
rate during the first few hundred decode steps after a tier transition is
substantially higher, because the working set has to rediscover its own
locality one block at a time.

\paragraph{Why clustering matters.} The 9--58\% throughput gain under memory
pressure reported in Section~\ref{sec:memory} depends on this mechanism.
Without PMI clustering the warm-up tail would be long enough that paging's
steady-state advantage would be amortized over a smaller useful window.
Clustering causes the working set to stabilize within the first few forward
passes rather than after several hundred.

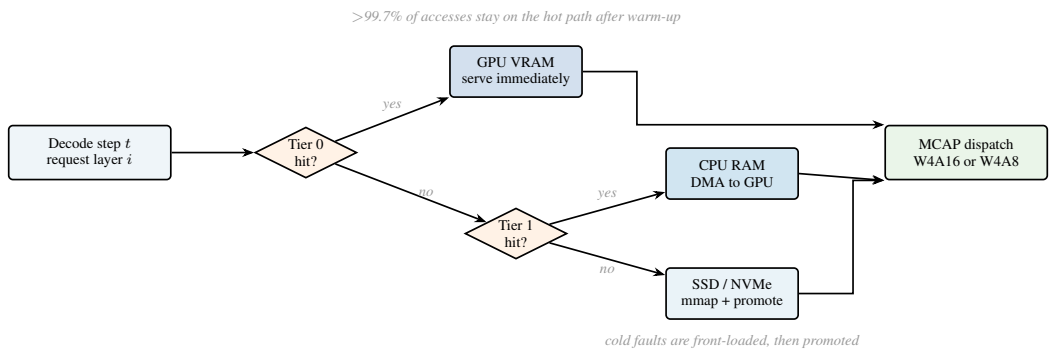
\begin{figure}[H]
\centering
\resizebox{\linewidth}{!}{
\begin{tikzpicture}[
    proc/.style={draw, rounded corners=2pt, minimum width=2.7cm, minimum height=0.9cm,
                 align=center, font=\scriptsize, thick, fill=nveblue!8},
    decision/.style={diamond, draw, aspect=2.0, align=center, font=\scriptsize, thick,
                     fill=nveorange!10, inner sep=1pt},
    tier/.style={draw, rounded corners=2pt, minimum width=2.2cm, minimum height=0.85cm,
                 align=center, font=\scriptsize, thick},
    arr/.style={-{Stealth[length=2mm]}, thick},
    note/.style={font=\scriptsize\itshape, text=nvegray},
]
\node[proc] (token) at (0,0) {Decode step $t$\\request layer $i$};
\node[decision] (hot) at (3.6,0) {Tier 0\\hit?};
\node[tier, fill=tier0!20] (tier0) at (7.1,1.35) {GPU VRAM\\serve immediately};
\node[decision] (warm) at (7.1,-1.35) {Tier 1\\hit?};
\node[tier, fill=tier1!30] (tier1) at (10.7,-0.35) {CPU RAM\\DMA to GPU};
\node[tier, fill=tier2!45] (tier2) at (10.7,-2.35) {SSD / NVMe\\mmap + promote};
\node[proc, fill=w4a8color!12] (dispatch) at (14.6,0) {MCAP dispatch\\W4A16 or W4A8};

\draw[arr] (token) -- (hot);
\draw[arr] (hot) -- node[above, note] {yes} (tier0);
\draw[arr] (hot) -- node[right, note] {no} (warm);
\draw[arr] (warm) -- node[above, note] {yes} (tier1);
\draw[arr] (warm) -- node[below, note] {no} (tier2);
\draw[arr] (tier0.east) -- ++(0.9,0) |- (dispatch.north west);
\draw[arr] (tier1.east) -- (dispatch.south west);
\draw[arr] (tier2.east) -- ++(0.9,0) |- (dispatch.south west);

\node[note] at (7.1,2.25) {$>$99.7\% of accesses stay on the hot path after warm-up};
\node[note] at (10.7,-3.15) {cold faults are front-loaded, then promoted};
\end{tikzpicture}
}
\caption{\textbf{Per-token paging lifecycle during decode.} Each layer access first checks
GPU residency, then CPU RAM, then SSD. Only the selected layer's weights move upward in the
hierarchy; once resident, the layer is dispatched to W4A16 or W4A8 based on its MCAP score.
The systems-level claim is about \emph{working-set stabilization}: expensive
cold faults are concentrated during warm-up, while steady-state decode
remains on the hot path.}
\label{fig:paging_lifecycle}
\end{figure}

\paragraph{Why the pager helps after warm-up.} Figure~\ref{fig:paging_lifecycle}
illustrates the steady-state mechanism. NVE does not stream the full model
from storage on every token. Instead it converges to a stable hot working set:
repeatedly accessed layers stay resident in Tier~0, warm layers are cheap to
re-promote from CPU RAM, and true SSD faults are paid mostly during initial
access. This is why the memory-constrained configuration shows no
observable quality loss in the evaluated regimes while sustaining
competitive decode throughput.

\begin{table}[H]
\centering
\caption{\textbf{Paging cache occupancy across 8 model scales.} Cold faults are paid
at most once per layer during warm-up (Faults $\leq$ Layers; the bound is tight for
models whose working set exceeds the hot tier throughout warm-up, loose when early
layers are already resident), after which steady-state decode is served from the hot tier. The $>$99.6\% ``hit rate'' is therefore a
statement about warm-up amortization, not about the quality of the replacement policy;
see Figure~\ref{fig:paging_lifecycle} for the lifecycle and Section~\ref{sec:memory} for the
operationally meaningful throughput claims.}
\label{tab:paging_stats}
\small
\begin{tabular}{lrrrr}
\toprule
Model & Layers & Hits & Faults & Hit Rate \\
\midrule
GPT-2 (0.1B) & 12 & 3,719 & 12 & 99.68\% \\
Qwen2.5 (0.5B) & 24 & 8,639 & 24 & 99.72\% \\
Llama-3.2 (1.2B) & 16 & 4,319 & 12 & 99.72\% \\
Qwen2 (1.5B) & 28 & 11,479 & 28 & 99.76\% \\
Llama-3.2 (3.2B) & 28 & 10,051 & 28 & 99.72\% \\
Qwen2 (3.1B) & 36 & 14,723 & 36 & 99.76\% \\
Llama-3.1 (8B) & 32 & 13,119 & 32 & 99.76\% \\
Qwen2 (7.6B) & 28 & 11,479 & 28 & 99.76\% \\
\bottomrule
\end{tabular}
\end{table}

\subsection{Architecture-Agnostic Weight Mapping}
\label{sec:arch}

\begin{figure}[H]
\centering
\resizebox{0.98\linewidth}{!}{%
\begin{tikzpicture}[
    shard/.style={draw, rounded corners=1pt, minimum width=1.1cm, minimum height=0.5cm,
                  font=\scriptsize, fill=nvegray!15, thick},
    stage/.style={draw, rounded corners=3pt, minimum width=3.2cm, minimum height=1.8cm,
                  font=\small, align=center, thick},
    slot/.style={draw, minimum width=1.0cm, minimum height=0.45cm,
                 font=\scriptsize, inner sep=1pt},
    arr/.style={-{Stealth[length=2.5mm]}, thick, nveblue},
    note/.style={font=\scriptsize\itshape, text=nvegray},
]
\node[shard] (s1) at (-0.2, 0.9) {\texttt{model-00001.st}};
\node[shard] (s2) at (0,   0.45) {\texttt{model-00002.st}};
\node[shard] (s3) at (0.2, 0.0)  {\texttt{model-00003.st}};
\node[shard] (s4) at (0.4, -0.45) {\texttt{\dots}};
\node[note, above=4pt of s1] {HuggingFace shards};
\node[note, below=2pt of s4] {(arch-specific names)};

\node[stage, fill=nvepurple!15] (norm) at (4, 0.2) {
  \textbf{Name Normalizer}\\[2pt]
  \scriptsize Llama $\cdot$ Qwen $\cdot$ Phi $\cdot$\\
  \scriptsize GPT-2 $\cdot$ GPT-NeoX $\cdot$ \ldots\\[2pt]
  \scriptsize \textit{split QKV, split gate+up,}\\
  \scriptsize \textit{transpose Conv1D}
};

\node[stage, fill=tier1!35] (ir) at (8.2, 0.2) {
  \textbf{\texttt{GenericBlockWeights}}\\[2pt]
  \scriptsize Canonical per-layer slots:\\[2pt]
};
\node[slot, fill=nvegray!10] at ($(ir.center)+(-1.0, -0.1)$) {\texttt{q}};
\node[slot, fill=nvegray!10] at ($(ir.center)+(0,    -0.1)$) {\texttt{k}};
\node[slot, fill=nvegray!10] at ($(ir.center)+(1.0,  -0.1)$) {\texttt{v}};
\node[slot, fill=nvegray!10] at ($(ir.center)+(-1.0, -0.55)$) {\texttt{o}};
\node[slot, fill=nvegray!10] at ($(ir.center)+(0,    -0.55)$) {\texttt{gate}};
\node[slot, fill=nvegray!10] at ($(ir.center)+(1.0,  -0.55)$) {\texttt{up/down}};

\node[stage, fill=w4a8color!15] (pack) at (12.6, 0.2) {
  \textbf{NVE Packed Layout}\\[2pt]
  \scriptsize 18-byte Q4\_0 blocks\\
  \scriptsize + INT8 per-group scales\\[2pt]
  \scriptsize \textit{consumed by 17}\\
  \scriptsize \textit{fused CUDA kernels}
};

\draw[arr] (s4.east)++(0.1,0.4) -- (norm.west);
\draw[arr] (norm.east) -- (ir.west);
\draw[arr] (ir.east)   -- (pack.west);

\node[font=\scriptsize\itshape, text=nvepurple] at (8.2, -1.35)
  {same IR consumed by MCAP profiler \& pager};
\draw[dashed, nvepurple] (ir.south) -- ++(0, -0.5);

\end{tikzpicture}%
}
\caption{\textbf{Architecture-agnostic weight mapping.} HuggingFace safetensors shards
across 12+ architectures flow through a single normalizer into a canonical
\texttt{GenericBlockWeights} IR, then into NVE's Q4\_0 packed layout. All downstream
components---MCAP profiler, weight pager, per-layer dispatch, and the 17 CUDA
kernels---operate on the IR, not on arch-specific names. One profile fits every target.}
\label{fig:weight_mapping}
\end{figure}
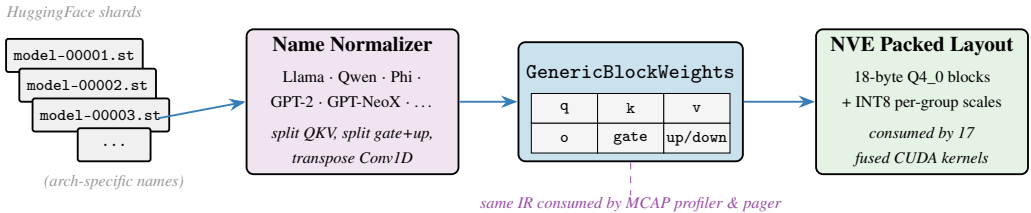

NVE supports 12+ model architectures through a generic weight mapping layer:
fused QKV projections (GPT-2, Phi-3, GPT-NeoX) are split into Q/K/V;
fused gate+up projections (Phi-3) are split into gate/up;
Conv1D weights (GPT-2) are transposed;
bias and no-bias architectures are handled transparently.
All profiling, paging, quantization, and dispatch logic operates on a unified
\texttt{GenericBlockWeights} abstraction (Figure~\ref{fig:weight_mapping}).

\subsection{Deployment Modes: A Spectrum Driven by MCAP}
\label{sec:deployment_modes}

The precision dispatch of Section~\ref{sec:w4a8_kernel} and the weight pager of
Section~\ref{sec:paging} are two points on a broader spectrum: given the same MCAP signal,
NVE exposes three execution modes that trade quality, memory, and simplicity
differently. A single command-line flag selects between them; the same profile and the
same kernels back all three.

\begin{description}[leftmargin=1.5em, itemsep=3pt]
\item[\textbf{Config A ,  Paged (baseline).}] All layers retained and tiered across
  GPU$\to$RAM$\to$SSD. MCAP drives initial placement; LRU + PMI-clustered prefetching
  maintain the hot working set (Section~\ref{sec:paging}). Quality is unchanged from the
  unconstrained model at any memory budget. This is the mode to pick when the
  compression ratio pushes the active-layer fraction below the hot-only viability floor
  (discussed below).

\item[\textbf{Config B ,  Hot-only.}] Layers below the MCAP threshold are
  \emph{skipped entirely} in the forward pass: not paged, not cached, not
  stored on device. The residual stream carries through the skipped positions.
  This is the most aggressive mode: no SSD needed, no PCIe traffic, no paging
  latency. Only the hot layers are loaded, and the rest never touch the device.
  The tradeoff is a bounded quality loss that depends on how many layers remain
  active.

\item[\textbf{Config C ,  Hot-only + AWQ.}] Hot layers additionally receive AWQ's
  saliency-weighted quantization. Within each retained layer, high-saliency channels get
  higher effective precision. On Llama-3.1-8B unconstrained this recovers factual-recall
  accuracy that the BF16 baseline loses (100\% vs.\ 88\% on the 8-task suite): AWQ's
  per-channel scaling protects the FFN key-value channels that encode factual
  associations. Config C is preferable in a narrow regime (3B and 8B
  unconstrained in our sweep); at 1B it fails, because the compression ratio
  drives the active-layer fraction below the hot-only viability floor
  (Table~\ref{tab:memory_constrained}). C is appropriate only when the post-AWQ
  layer budget still clears that floor.
\end{description}

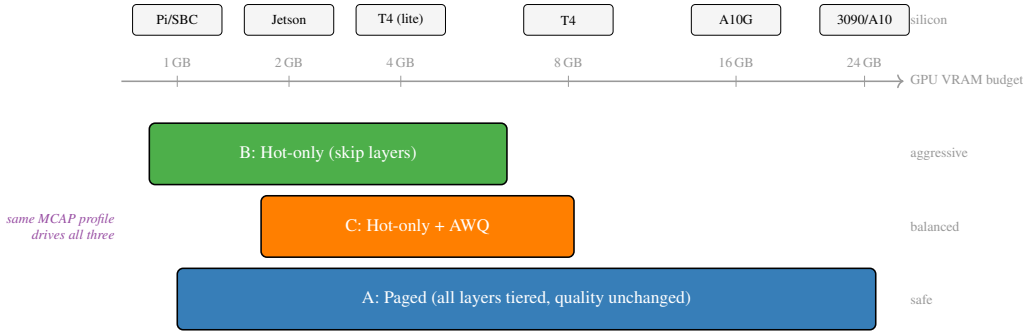
\begin{figure}[H]
\centering
\resizebox{0.98\linewidth}{!}{%
\begin{tikzpicture}[
    band/.style={draw, rounded corners=3pt, thick, minimum height=1.1cm,
                 font=\small, align=center, text=white},
    silicon/.style={draw, rounded corners=2pt, minimum width=1.6cm, minimum height=0.55cm,
                    font=\scriptsize, fill=nvegray!10},
    axlabel/.style={font=\scriptsize, text=nvegray},
    tick/.style={font=\scriptsize, text=nvegray},
]
\node[silicon] at (1,   1.6) {Pi/SBC};
\node[silicon] at (3,   1.6) {Jetson};
\node[silicon] at (5,   1.6) {T4 (lite)};
\node[silicon] at (8,   1.6) {T4};
\node[silicon] at (11,  1.6) {A10G};
\node[silicon] at (13.3, 1.6) {3090/A10};
\node[axlabel, anchor=west] at (14, 1.6) {silicon};

\draw[->, thick, nvegray] (0, 0.5) -- (14, 0.5);
\node[axlabel, anchor=west] at (14, 0.5) {GPU VRAM budget};
\foreach \x/\lbl in {1/1\,GB, 3/2\,GB, 5/4\,GB, 8/8\,GB, 11/16\,GB, 13.3/24\,GB} {
  \draw[nvegray, thin] (\x, 0.4) -- (\x, 0.6);
  \node[tick, above=1pt] at (\x, 0.6) {\lbl};
}

\node[band, fill=nvegreen, minimum width=6.4cm] at (3.7, -0.8) {B: Hot-only (skip layers)};
\node[band, fill=nveorange, minimum width=5.6cm] at (5.3, -2.1) {C: Hot-only + AWQ};
\node[band, fill=nveblue, minimum width=12.5cm] at (7.25, -3.4) {A: Paged (all layers tiered, quality unchanged)};

\node[axlabel, anchor=west] at (14, -0.8) {aggressive};
\node[axlabel, anchor=west] at (14, -2.1) {balanced};
\node[axlabel, anchor=west] at (14, -3.4) {safe};

\node[font=\scriptsize\itshape, text=nvepurple, anchor=east, align=right] at (0, -2.1)
  {same MCAP profile\\drives all three};

\end{tikzpicture}%
}
\caption{\textbf{Deployment spectrum driven by a single MCAP profile.} The same
60-second profile selects between three execution modes. Hot-only (B) fits the tightest
budgets by skipping low-importance layers but is bounded by the $\sim$50\,\% active-layer
floor; Hot+AWQ (C) applies saliency-weighted quantization to the retained layers for
additional headroom; Paged (A) tiers all layers across GPU$\to$RAM$\to$SSD with quality
unchanged and is the correct choice when the compression ratio would otherwise push B/C
below the floor. Bands show approximate viable budget ranges; silicon targets indicate
the budgets where each mode typically dominates.}
\label{fig:deployment_spectrum}
\end{figure}

\paragraph{The hot-only viability floor.} Configs B and C skip layers entirely, so they
are bounded by how many layers the residual stream can tolerate losing
(Figure~\ref{fig:deployment_spectrum} situates the three modes on the GPU-budget axis). We observe an
\emph{empirical cliff} near 50\,\% active layers across the three sweep points we ran
(25\,\%, 36\,\%, 50\,\% active): above the cliff, skipped layers contribute
near-redundant updates and output remains coherent; below it, output quality collapses.
The transition width is not resolved by the current sweep, and the 50\,\% figure is a
curve-fit to a small grid rather than a derived quantity;
Table~\ref{tab:deployment_modes_summary} summarizes the observed behavior.

\begin{table}[H]
\centering
\caption{\textbf{Deployment mode viability by model and memory budget.} ``Active'' is
the fraction of layers retained in the GPU hot tier. Hot-only modes (B, C) fail below
roughly 50\% active; paging (A) is the fallback for aggressive compression
ratios. Throughput numbers are ABC-suite measurements on T4 (pre-W4A8 runs;
the current W4A8 path is strictly faster).}
\label{tab:deployment_modes_summary}
\small
\begin{tabular}{llcccl}
\toprule
Model & Budget & Active & A (paged) & B (hot) & C (hot+AWQ) \\
\midrule
3B & unconstrained  & 100\% & 100\%  & 100\%  & 100\% \\
3B & 2\,GB          & 36\%  & \textbf{100\%} & 0\% (floor) & 0\% (floor) \\
8B & unconstrained  & 100\% & 88\%   & 88\%   & \textbf{100\%} \\
8B & 8\,GB          & 50\%  & \textbf{88\%}  & 12\% (at floor) & 0\% \\
8B & 4\,GB          & 25\%  & \textbf{88\%}  & 0\% (floor) & 0\% \\
1B & any            & 100\% & 88\%   & 88\%   & 0\% (bpw floor\,$^\dagger$) \\
\bottomrule
\end{tabular}

\smallskip
{\footnotesize $^\dagger$Config C's 2.0\,bpw target is too aggressive for a 16-layer
model: the quantization-depth floor is reached before the layer-count floor.
We report this failure mode explicitly.}
\end{table}

\paragraph{Decision rule.} MCAP scores plus a target memory budget determine the mode:
\begin{enumerate}[leftmargin=*, itemsep=1pt]
\item If the budget keeps $\geq$50\% of layers active on GPU, prefer \textbf{Config C}
  (hot-only + AWQ): fastest, lowest footprint, best factual recall in the tested 8B
  regime. Fall back to \textbf{Config B} if AWQ is degenerate for the model (as on 1B).
\item If the budget pushes below the active-layer floor, switch to \textbf{Config A}
  (paging): no observable quality loss in the evaluated regimes at arbitrary memory ratios, with PMI-clustered
  prefetching keeping steady-state throughput competitive.
\item In both cases, \textbf{precision dispatch} (W4A8 for sub-threshold layers, W4A16
  for outliers) applies \emph{within} the mode, hot-only and paged configurations both
  route per-layer precision on the same MCAP signal.
\end{enumerate}

\paragraph{Implications for memory-constrained targets.} The hot-only modes
are what make sub-$\sim$4\,GB operation plausible without an SSD in the loop.
Config B/C on Llama-3.2-3B with 14 of 28 layers retained (50\% active, at the
floor) runs entirely in GPU VRAM with zero paging traffic, a regime relevant to
hardware where PCIe to CPU RAM is slow and SSD storage is limited or absent.
Config A extends the envelope further when the memory budget pushes below the
floor, at the cost of paging overhead. Together, the three modes span a range
from server-class GPUs down to memory-constrained embedded devices, driven by
one profile.

\section{Experiments}
\label{sec:experiments}

\subsection{Setup}

\paragraph{Hardware.} GPU throughput experiments use NVIDIA T4 (16\,GB VRAM, sm\_75) on
Modal cloud. Quality evaluations additionally run on A10G (24\,GB VRAM). All results are
publicly reproducible with a free Modal account.

\paragraph{Models.} End-to-end throughput and deployment experiments focus on
Llama-3.2-1B (16 layers), Llama-3.2-3B (28 layers), and Llama-3.1-8B (32 layers). MCAP
profiling, scorer analysis, and paging behavior are validated more broadly across eight
model scales from GPT-2 (0.1B) through Qwen2-7.6B and Llama-3.1-8B
(Table~\ref{tab:mcap_scores}).

\paragraph{Baselines.} NVE W4A16 (uniform), NVE W4A8 (uniform), NVE MCAP Mixed
(threshold $\tau = 0.7$), llama.cpp Q4\_0 with ggml-cuda backend, AutoAWQ INT4
(group\_size=128), and HuggingFace FP16.

\paragraph{Evaluation.}
WikiText-2 perplexity (1B: 50 sequences, 3B: 10 sequences; 256 tokens each);
HellaSwag accuracy (1B: $n=50$, 3B: $n=20$; 4-way log-likelihood);
8-task generative suite (2 QA, 2 reasoning, 2 coding, 2 summarization prompts; tasks
fixed before any MCAP experiment was run and held constant across all configurations);
BenchRandom decode throughput (2,000 iterations, real weights). The tightest
matched-quality comparisons are reported for selected 1B/3B regimes; throughput
and memory-constrained operation are evaluated through 8B.

\paragraph{Statistical caveat.} At $n=8$ the 95\,\% binomial Wilson interval around a
point estimate of $7/8 = 87.5\,\%$ is $[52.9, 97.8]$\,\%, so single-row accuracy
differences on the generative suite are not hypothesis tests; we treat
same-point-estimate rows as ``indistinguishable at the suite's resolution'' rather than
equal. WikiText-2 (12,800 tokens at 50 seq.) and HellaSwag ($n=50$) carry tighter
resolution and are the primary quality evidence; the 8-task suite is a coarse generative
consistency check rather than a benchmark.

\subsection{Kernel Throughput: 1.5--1.8$\times$ Over llama.cpp}
\label{sec:throughput}

\begin{table}[H]
\centering
\caption{\textbf{Decode throughput (tok/s).} BenchRandom, real weights, T4
GPU, batch=1, 2,000 iterations. NVE W4A8 exceeds llama.cpp Q4\_0 by
1.5--1.8$\times$ across all scales. The advantage over W4A16 grows with model
size, consistent with a larger fraction of bandwidth-bound matmuls at scale.}
\label{tab:throughput}
\small
\begin{tabular}{lrrrrr}
\toprule
Model & NVE W4A8 & NVE W4A16 & llama.cpp & vs W4A16 & vs llama.cpp \\
\midrule
Llama-3.2-1B & \textbf{269.1} & 116.7 & 150.8 & 2.31$\times$ & \textbf{1.78$\times$} \\
Llama-3.2-3B & \textbf{108.8} & 43.0 & 70.9 & 2.53$\times$ & \textbf{1.53$\times$} \\
Llama-3.1-8B & \textbf{47.7} & 16.7 & 30.8 & 2.86$\times$ & \textbf{1.55$\times$} \\
\bottomrule
\end{tabular}
\end{table}

\begin{figure}[H]
\centering
\begin{tikzpicture}
\begin{axis}[
    width=\linewidth, height=6cm,
    ybar,
    bar width=12pt,
    xlabel={Model},
    ylabel={Throughput (tok/s)},
    symbolic x coords={1B, 3B, 8B},
    xtick=data,
    legend style={at={(0.98,0.98)}, anchor=north east, font=\small},
    ymin=0, ymax=310,
    grid=major,
    grid style={gray!20},
    nodes near coords,
    nodes near coords style={font=\tiny, above},
]
\addplot[fill=w4a8color!80, draw=w4a8color] coordinates {(1B, 269.1) (3B, 108.8) (8B, 47.7)};
\addplot[fill=w4a16color!80, draw=w4a16color] coordinates {(1B, 116.7) (3B, 43.0) (8B, 16.7)};
\addplot[fill=llamacppcolor!80, draw=llamacppcolor] coordinates {(1B, 150.8) (3B, 70.9) (8B, 30.8)};
\legend{NVE W4A8, NVE W4A16, llama.cpp Q4\_0}
\end{axis}
\end{tikzpicture}
\caption{\textbf{Decode throughput comparison.} NVE W4A8 exceeds llama.cpp
Q4\_0 by 1.5--1.8$\times$. The W4A8 advantage over W4A16 grows with model
size: 2.31$\times$ at 1B, 2.53$\times$ at 3B, 2.86$\times$ at 8B. The
three-point curve shows the scaling trend directly rather than relying on
isolated benchmarks.}
\label{fig:throughput_bars}
\end{figure}
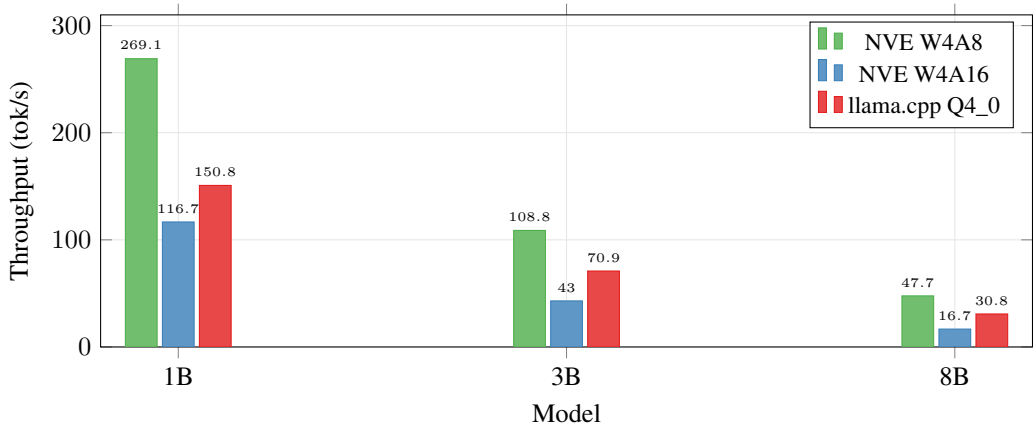

\paragraph{Why the speedup scales with model size.} Larger models have proportionally more
bandwidth-bound matmuls. W4A8 reduces activation data movement by 2$\times$ (INT8 vs.\ F16),
and this bandwidth savings translates more directly to throughput when operations are
memory-bound. At 1B, some projections are partially compute-bound on T4; at 8B, the dominant
FFN projections (14336$\times$4096) are fully bandwidth-limited.

\paragraph{Cost implications.} Under T4 pricing of \$0.35/GPU-hour, the
throughput difference corresponds to \$0.36/M tokens (1B) vs.\ llama.cpp's
\$0.64/M, a 44\% reduction at matched quality; at 8B, \$2.04/M vs.\ \$3.15/M
(35\% reduction). The gap grows with model size because the throughput gap
does. Figure~\ref{fig:throughput_bars} shows that the speedup is not a
small-model effect: the same W4A8 kernel remains favorable at 3B and 8B, and
the gap over W4A16 widens as the workload becomes more bandwidth-bound.

\paragraph{Kernel evolution.} Table~\ref{tab:kernel_evolution} traces the throughput
progression from unfused operations to the final W4A8 system:

\begin{table}[H]
\centering
\caption{\textbf{Kernel optimization progression (Llama-3.2-1B, T4).} The
W4A8 dp4a kernel provides the single largest improvement, approximately
doubling throughput and crossing the llama.cpp Q4\_0 baseline.}
\label{tab:kernel_evolution}
\small
\begin{tabular}{lrrr}
\toprule
Configuration & tok/s & ms/tok & vs llama.cpp \\
\midrule
Unfused F16 baseline & 92.2 & 10.84 & 0.61$\times$ \\
+ Fused RMSNorm + RoPE & 109.3 & 9.15 & 0.72$\times$ \\
+ Fused Matvec (W4A16) & 114.9 & 8.70 & 0.76$\times$ \\
+ W4A16 quantized matvec & 126.4 & 7.91 & 0.84$\times$ \\
+ Flash decode + fused QKV & 124.1 & 8.06 & 0.82$\times$ \\
\rowcolor{w4a8color!15}
+ \textbf{W4A8 dp4a kernel} & \textbf{269.1} & \textbf{3.72} & \textbf{1.78$\times$} \\
\bottomrule
\end{tabular}
\end{table}

\subsection{Quality in the Tested Regimes}
\label{sec:quality}

\begin{table}[H]
\centering
\caption{\textbf{WikiText-2 perplexity (A10G).} $\Delta \leq 0.01$ confirms negligible
degradation from INT8 activation quantization. Lower is better. Sample sizes differ: 1B
is a 50-sequence run, 3B is a 10-sequence run from the same evaluation harness.}
\label{tab:wikitext}
\small
\begin{tabular}{lccccc}
\toprule
Model & W4A16 & W4A8 & MCAP Mixed & $\Delta$ & Sequences \\
\midrule
Llama-3.2-1B & 17.51 & 17.51 & \textbf{17.51} & 0.00 & 50 \\
Llama-3.2-3B & 14.01 & 14.01 & \textbf{14.01} & 0.00 & 10 \\
\bottomrule
\end{tabular}
\end{table}

\begin{table}[H]
\centering
\caption{\textbf{HellaSwag accuracy} (4-way log-likelihood, A10G). All strategies produce
identical results. The 1B run uses $n$=50; the 3B run uses $n$=20 from the same harness.
Small $n$ means the accuracy numbers carry meaningful binomial variance (1B: $\pm$6.9 pp
at 95\%; 3B: $\pm$10.5 pp), so these tables support an \emph{equivalence} claim across
precision strategies, not absolute benchmark placement.}
\label{tab:hellaswag}
\small
\begin{tabular}{lcccc}
\toprule
Model & W4A16 & W4A8 & MCAP Mixed & Correct/Total \\
\midrule
Llama-3.2-1B & 54.0\% & 54.0\% & 54.0\% & 27/50 \\
Llama-3.2-3B & 65.0\% & 65.0\% & 65.0\% & 13/20 \\
\bottomrule
\end{tabular}
\end{table}

\begin{table}[H]
\centering
\caption{\textbf{8-task generative accuracy} (QA, reasoning, coding, summarization).
Matched accuracy in the tested configurations.}
\label{tab:task_acc}
\small
\begin{tabular}{lcccc}
\toprule
Model & W4A16 & W4A8 & MCAP Mixed & $n$ \\
\midrule
Llama-3.2-1B & 88\% & 88\% & 88\% & 8 \\
Llama-3.2-3B & 100\% & 100\% & 100\% & 8 \\
Llama-3.1-8B & 100\% & 100\% & --- & 8 \\
\bottomrule
\end{tabular}
\end{table}

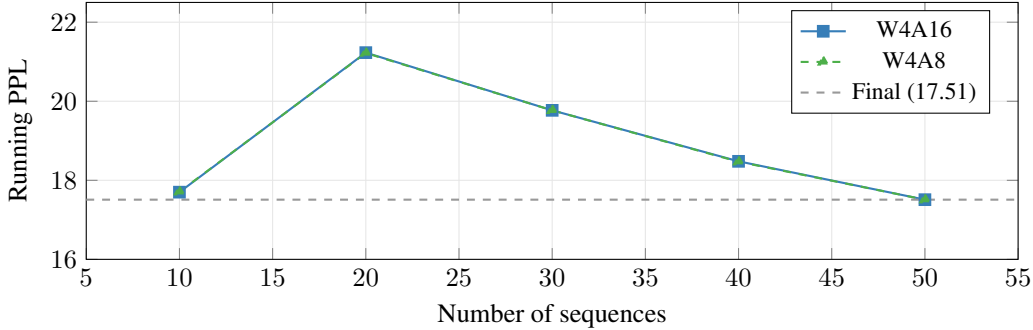
\begin{figure}[H]
\centering
\begin{tikzpicture}
\begin{axis}[
    width=\linewidth, height=5cm,
    xlabel={Number of sequences},
    ylabel={Running PPL},
    legend pos=north east,
    legend style={font=\small},
    grid=major,
    grid style={gray!20},
    xmin=5, xmax=55,
    ymin=16, ymax=22.5,
]
\addplot[thick, w4a16color, mark=square*, mark size=2] coordinates {
    (10, 17.70) (20, 21.23) (30, 19.77) (40, 18.48) (50, 17.51)
};
\addlegendentry{W4A16}
\addplot[thick, w4a8color, mark=triangle*, mark size=2, dashed] coordinates {
    (10, 17.71) (20, 21.22) (30, 19.77) (40, 18.47) (50, 17.51)
};
\addlegendentry{W4A8}
\addplot[thick, nvegray, dashed, no marks] coordinates {
    (5, 17.51) (55, 17.51)
};
\addlegendentry{Final (17.51)}
\end{axis}
\end{tikzpicture}
\caption{\textbf{WikiText-2 running PPL convergence (Llama-3.2-1B, 50 sequences).}
W4A16 and W4A8 curves are visually indistinguishable. $\Delta \leq 0.01$ at every
checkpoint confirms zero systematic bias.}
\label{fig:ppl_convergence}
\end{figure}

For the 1B run, the matched-quality result is not an artifact of sequence
selection. Figure~\ref{fig:ppl_convergence} shows the running PPL convergence
across 50 sequences on Llama-3.2-1B: $\Delta \leq 0.01$ at every 10-sequence
checkpoint, converging to $\Delta = 0.00$ at 50 sequences. The 3B run is a
shorter 10-sequence evaluation from the same harness (no sequence-level
convergence plot), and should be read as directionally consistent evidence
rather than a 50-sequence validation. The larger-model analysis in this paper
is primarily systems-oriented---throughput scaling, paging behavior, and the
reachable memory budget through 8B---and the matched-quality claim itself
remains specific to the 1B and 3B setups explicitly evaluated here.

\subsection{Comparison with Existing PTQ Methods}
\label{sec:ptq_comparison}

Table~\ref{tab:ptq} compares NVE against AutoAWQ~\cite{lin2023awq} at matched 4-bit
precision, with HuggingFace FP16 as the quality ceiling.

\begin{table}[H]
\centering
\caption{\textbf{Comparison with AutoAWQ (Llama-3.2-1B, A10G).} WikiText-2 PPL
(50 seq $\times$ 256 tok), HellaSwag ($n$=50). AWQ achieves lower PPL via per-channel
saliency weighting, while NVE achieves higher downstream accuracy with
no offline calibration and no weight modification. The two operate at
different granularities (within-layer vs.\ across-layer) and can be
composed.}
\label{tab:ptq}
\small
\begin{tabular}{lcccc}
\toprule
Method & Calibration & PPL ($\downarrow$) & HellaSwag ($\uparrow$) & Weights modified? \\
\midrule
HF FP16 (ceiling) & --- (no quant) & \textbf{14.94} & \textbf{56.0\%} & No \\
\midrule
AutoAWQ INT4 & 128 samples, offline & 16.64 & 52.0\% & Yes \\
\rowcolor{w4a8color!15}
NVE MCAP Mixed & 12 prompts, 60s & 17.51 & \textbf{54.0\%} & No \\
NVE W4A8 (uniform) & 12 prompts, 60s & 17.51 & 54.0\% & No \\
NVE W4A16 & --- (no calib.) & 17.51 & 54.0\% & No \\
\bottomrule
\end{tabular}
\end{table}

\paragraph{AWQ and MCAP operate at different granularities.} AWQ
achieves lower PPL (16.64 vs.\ 17.51) because per-channel saliency
weighting preserves high-variance weight channels during quantization.
NVE scores higher on HellaSwag (54.0\% vs.\ 52.0\%), indicating that PPL
and downstream accuracy do not always correlate at 4-bit precision. AWQ
operates within each layer (per-channel); MCAP operates across layers
(per-layer dispatch). The two compose directly: MCAP can route
AWQ-quantized weights through the W4A8 kernel for low-importance layers,
yielding lower PPL and higher throughput simultaneously. This works on
the 8B Config~C result reported later but fails at 1B, where the
compression ratio crosses the hot-only viability floor, so we report it
as a useful extension in the regime where the floor is cleared rather
than as a default.

\paragraph{Observed failure mode.} The brittleness is sharp rather than
gradual. In the 1B raw generations, aggressive profile-guided quantization
produces repetitive boilerplate and malformed continuations rather than
slightly worse answers. This distinguishes a stable operating region from an
unstable one. The positive claim is therefore not that every profile-guided
quantized configuration works, but that runtime importance information helps
identify which precision reductions preserve baseline behavior and which
cross a clear failure boundary.

\subsection{Ablation: Threshold Sensitivity}
\label{sec:threshold}

\begin{table}[H]
\centering
\caption{\textbf{Threshold ablation (Llama-3.2-1B).} 20 sequences $\times$ 256 tokens.
All configurations within $\pm$0.04 PPL. The threshold governs throughput, not quality.}
\label{tab:threshold}
\small
\begin{tabular}{lccc}
\toprule
Threshold & W4A16 layers & W4A8 layers & PPL \\
\midrule
all W4A8 ($\tau = 2.0$) & 0 / 16 & 16 / 16 & 21.22 \\
$\tau = 0.30$ & 5 / 16 & 11 / 16 & 21.21 \\
$\tau = 0.10$ & 10 / 16 & 6 / 16 & 21.24 \\
$\tau = 0.05$ & 14 / 16 & 2 / 16 & 21.24 \\
all W4A16 ($\tau = 0.0$) & 16 / 16 & 0 / 16 & 21.23 \\
\bottomrule
\end{tabular}
\end{table}

The $\pm$0.04 PPL spread indicates that W4A8 introduces negligible noise
regardless of how many layers use it. The threshold controls throughput
rather than quality. The 50-sequence evaluation
(Table~\ref{tab:wikitext}) confirms that all strategies converge to
PPL=17.51 with sufficient data.

\subsection{Memory-Constrained Inference}
\label{sec:memory}

\begin{figure}[H]
  \centering
  \includegraphics[width=\linewidth]{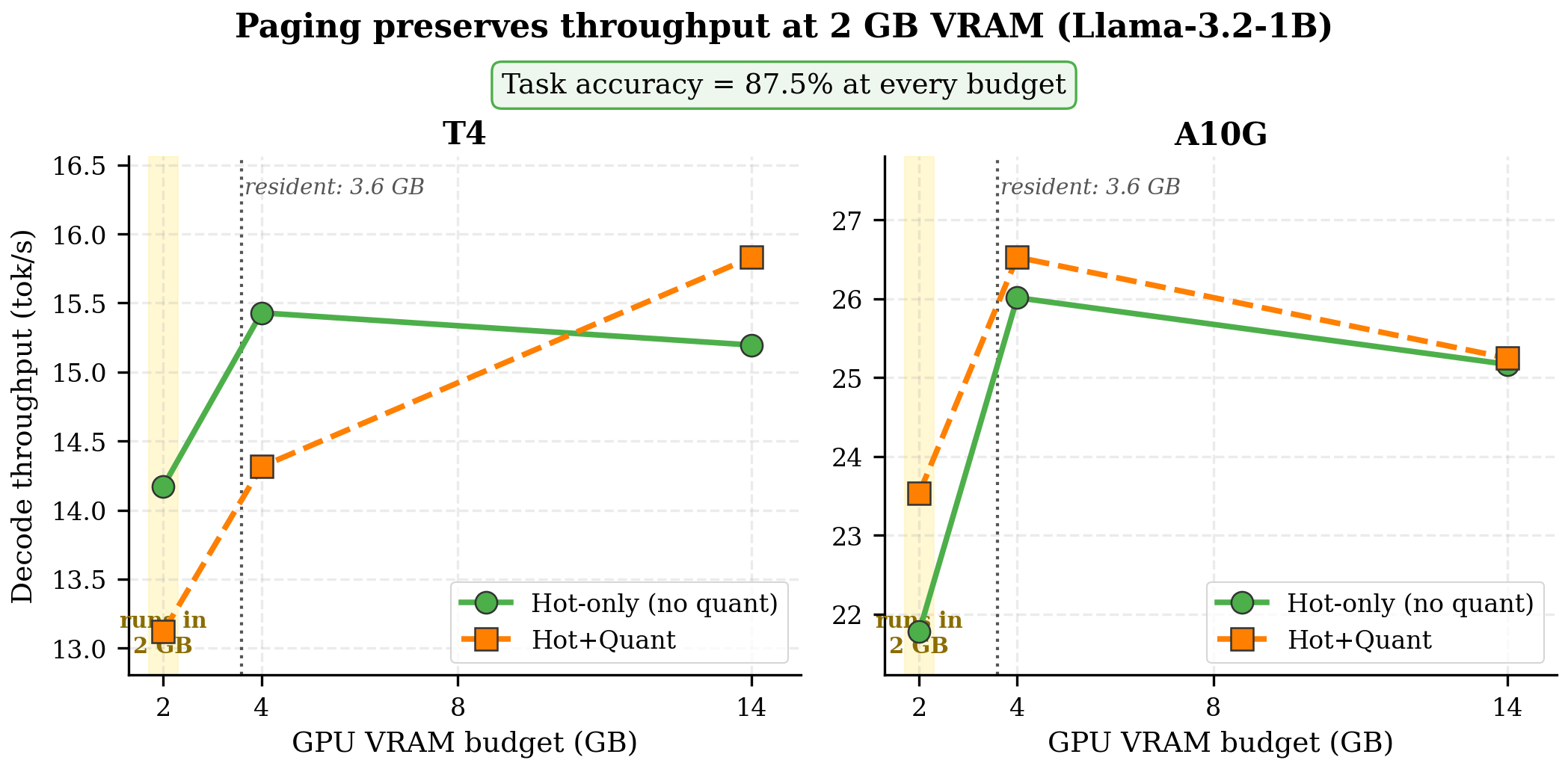}
  \caption{\textbf{Paging runs Llama-3.2-1B (unconstrained resident $\sim$3.6\,GB) at a 2\,GB GPU budget, quality intact.}
  Llama-3.2-1B decode throughput as the GPU VRAM budget is swept from 2\,GB to 14\,GB,
  on T4 and A10G. Hot-only paging (no quantization) matches Hot+Quant throughput at
  every budget, and task accuracy is 87.5\% across all runs---i.e.\ the model's
  full-precision weights exceed the smallest budget, yet paging maintains throughput
  and quality without any loss from quantization.}
  \label{fig:budget_sweep}
\end{figure}

\emph{Reading guide.} This section reports on a $\{1B, 3B, 8B\} \times \{$unconstrained,
2\,GB, 4\,GB, 8\,GB$\}$ sweep across three systems (NVE, llama.cpp, HF). The main
numbers are in Table~\ref{tab:memory_constrained}; the full A/B/C $\times$ scenario
matrix is in Table~\ref{tab:abc_matrix}; the hot-only floor at 8\,B is in
Table~\ref{tab:coherence_floor}; the single-profile cross-silicon demonstration is in
Figure~\ref{fig:budget_sweep}. The paragraphs below expand on each in that order.

The 3-tier pager (Section~\ref{sec:paging}) lets NVE serve models far below their raw weight
footprint. For reference: Llama-3.2-3B needs $\sim$6.4\,GB of BF16 weights, Llama-3.1-8B
needs $\sim$16\,GB, and even at 4-bit an 8B model still requires $\sim$5--6\,GB
resident. We evaluate the pager across three budgets per model scale:
unconstrained (full GPU), moderate (8\,GB for 8B), and tight (2\,GB for 3B,
4\,GB for 8B). The question we ask: does paging cost accuracy?

In the tested configurations, no. Table~\ref{tab:memory_constrained} reports
the main result: NVE runs Llama-3.2-3B with full BF16 weights in a 2\,GB
budget, 3.5$\times$ below its weight footprint, with no observable
degradation on the evaluated task suite relative to the unconstrained run. At 8B, NVE operates in 4\,GB, below the
$\sim$5\,GB floor that standard 4-bit quantization requires. llama.cpp and
HuggingFace both OOM in these configurations. NVE remains operational by
keeping only the MCAP-identified hot layers GPU-resident and paging the rest
through CPU RAM and SSD at a $>$99.7\% hit rate
(Table~\ref{tab:paging_stats}).

\paragraph{Why the constrained baseline works.} The most load-bearing result
in this section is the constrained \emph{baseline}, not the more aggressive
profiled variants. The baseline result indicates that NVE is not simply
another quantization recipe but a weight-virtualization layer that, in the
evaluated settings, shows no observable degradation from full-precision
behavior while operating under memory budgets that appear infeasible from
the raw weight size alone. The pager works
because decode repeatedly revisits the same small set of important layers;
once these layers are resident, later decode steps hit in the hot tier, and
the observed $>$99.7\% hit rates indicate that paging cost is concentrated
during warm-up rather than paid on every token.

A less obvious finding is that, on the paged BF16 path, constraining memory
does not slow decode down and can speed it up. Comparing the paged BF16
baseline across scenarios (Table~\ref{tab:memory_constrained}): Llama-3.2-1B
moves from 3.22 tok/s unconstrained to 4.23 tok/s at 2\,GB (+31\%);
Llama-3.2-3B from 2.12 to 2.31 tok/s (+9\%); and Llama-3.1-8B from 0.79 to
1.25 tok/s at 4\,GB (+58\%), with unchanged task accuracy. The mechanism is
working-set reduction: MCAP-guided paging keeps only the high-importance
layers GPU-resident, which reduces memory pressure and improves cache
locality; cold layers are accessed only during initial warm-up, after which
steady-state decode stays on the hot path.

\paragraph{Scope of the throughput numbers in Table~\ref{tab:memory_constrained}.} These
are \emph{paged BF16} decode throughputs, not the W4A8 dp4a kernel throughputs reported
in Table~\ref{tab:throughput}. The two columns are deliberately different paths: the
kernel table measures the fast W4A8 GPU-resident path (269 tok/s on 1B), whereas the
memory-constrained table measures the pager end-to-end with full-precision weights moved
across the tier hierarchy. Both are end-to-end paths in NVE; the kernel path
addresses throughput, and the paging path addresses feasibility. Conflating
them would overstate either.

\paragraph{Anomalous 8B 8\,GB point.} Table~\ref{tab:memory_constrained}'s 8B row at
8\,GB reports 0.55 tok/s, lower than both the unconstrained 0.79 and the tighter 4\,GB
1.25. This is a real measurement from our memory-constrained benchmark sweep, not a
reporting error, and it runs against the monotonic trend the other rows suggest. We
offer a tentative explanation: at 8\,GB the hot budget is large enough to admit many
lower-importance layers that end up thrashing against the KV cache and transient
activation storage, while 4\,GB forces a sharper hot-cold partition that MCAP handles
well. We have not isolated the cause in a controlled ablation and report this
openly. The result should not be generalized as ``more constraint is always
faster''; the correct reading is that working-set-reduction effects can
dominate tier-transfer costs at a well-chosen partition, while a poorly chosen
partition can be worse than either extreme.

\paragraph{Beyond GPUs: CPU-only operation.} The pager is GPU-agnostic. On a
commodity CPU-only machine (3.8\,GB RAM, 2 AMD vCPUs, no GPU), NVE runs
Qwen2.5-0.5B end-to-end at 6.9 tok/s decode with a 512\,MB hot tier, 171\,MB
warm tier, and 96\,MB KV cache ($\sim$779\,MB total resident footprint) at
zero page faults (Appendix~\ref{app:reproducibility}). This is outside the
primary regime the paper's throughput claims target, but it shows that the
three-tier abstraction degrades gracefully to environments that llama.cpp and
similar systems already serve, while remaining the only path we are aware of
that supports full-BF16 multi-GB-model execution on sub-GB GPUs.

\paragraph{Mechanistic interpretation.} The throughput gain should not be read as claiming
that PCIe or SSD are faster than VRAM. The relevant effect is \emph{working-set reduction}.
Under an unconstrained placement, too many weights compete for GPU residency and cache space.
Under NVE's constrained placement, only the repeatedly useful portion of the model remains
resident in the hot tier, while infrequently used weights are demoted after initial access.
If the working set is chosen well, the system trades a small one-time migration cost for a
more stable steady state during autoregressive decode. The constrained 3B and 8B baseline
results suggest this is exactly what happens in the tested regimes.

At 8B scale in the unconstrained setting, Config~C (MCAP + AWQ weights) reaches 100\%
task accuracy on the tested suite, compared with 88\% for the BF16 baseline. We do not
interpret this isolated result as evidence that quantization is generally superior to
full-precision inference. A more conservative interpretation is that MCAP+AWQ can remain
competitive in selected regimes, and that small evaluation suites can expose variance in
task-level outcomes that should be validated at larger scale.

Figure~\ref{fig:budget_sweep} illustrates the consequence. Llama-3.2-1B,
whose full-precision weights occupy $\sim$2.5\,GB (and whose unconstrained
resident footprint, including KV cache, attention scratch, and CUDA-graph
buffers, is $\sim$3.6\,GB), runs at a 2\,GB GPU VRAM budget with decode
throughput and task accuracy unchanged from the 14\,GB-budget run. Hot-only
paging, without any quantization, matches the throughput of the
paging+quantization configuration on both T4 and A10G, and task accuracy is
87.5\% across every budget. A practical consequence is that commodity
hardware (e.g.\ laptops with 4\,GB discrete GPUs, embedded devices with
limited VRAM) can run larger models without losing the quality of those
models.
The larger-model operating points reported in Table~\ref{tab:task_acc} and
Figure~\ref{fig:abc_quality_throughput} show that this behavior persists at the scales
where VRAM pressure is operationally decisive, not only in smaller exploratory models.
Table~\ref{tab:coherence_floor} makes the negative boundary explicit. The failure mode is
not that profiling is unhelpful, but that \emph{hot-only} execution stops being a safe
substitute for full-layer inference once the active-layer ratio falls to roughly half of
the network or lower. In those regimes, paging all layers is the correct strategy.

\begin{table}[H]
\centering
\caption{\textbf{Memory-constrained inference: accuracy and throughput (paged BF16 path).}
Throughput numbers here are from the paged BF16 decode path, distinct from the W4A8
kernel path of Table~\ref{tab:throughput}. NVE serves 3B with full BF16 weights
($\sim$7\,GB) in a 2\,GB budget, and 8B ($\sim$16\,GB BF16) in 4\,GB, 3.5--4$\times$
below the weight footprint, with no task-accuracy penalty. Under constraint, working-set
reduction can \emph{raise} throughput (e.g., 3B 2\,GB: +9\%; 8B 4\,GB: +58\%), though the
8B 8\,GB row is an anomaly we flag rather than explain (see text). In the unconstrained
8B setting, Config C (MCAP + AWQ) attains 100\% on $n$=8 tasks vs.\ 88\% baseline; the
$n$=8 suite makes this a suggestive outlier, not an ordering claim.}
\label{tab:memory_constrained}
\small
\begin{tabular}{llcccc}
\toprule
Model & Scenario & NVE Base Acc & NVE Base tok/s & llama.cpp & HF FP16 \\
\midrule
1B & unconstrained & 88\% & 3.22 & 88\% & 100\% \\
1B & 2\,GB & 88\% & \textbf{4.23} (+31\%) & --- & --- \\
\midrule
3B & unconstrained & 100\% & 2.12 & 100\% & --- \\
\rowcolor{w4a8color!15}
3B & \textbf{2\,GB} & \textbf{100\%} & \textbf{2.31} (+9\%) & OOM & OOM \\
\midrule
8B & unconstrained & 88\% & 0.79 & 100\% & 88\% \\
\rowcolor{w4a8color!15}
8B & \textbf{4\,GB} & \textbf{88\%} & \textbf{1.25} (+58\%) & OOM & OOM \\
8B & 8\,GB & 88\% & 0.55 & 100\% & OOM \\
\bottomrule
\end{tabular}
\end{table}

\begin{table}[H]
\centering
\caption{\textbf{How to read the memory result.} The strongest operating points are the
constrained baseline rows: full-BF16 behavior maintained, with no observable
degradation on the evaluated suite, under budgets well below the raw
weight footprint. More aggressive profiled variants can be faster, but they are not the
main robustness claim.}
\label{tab:memory_takeaway}
\footnotesize
\begin{tabular}{llll}
\toprule
Model & Budget & Supported claim & Evidence in this paper \\
\midrule
Llama-3.2-3B & 2\,GB & Full-BF16 baseline behavior preserved & 100\% acc., 2.31 tok/s \\
Llama-3.1-8B & 4\,GB & Baseline behavior matched under tight budget & 88\% acc., 1.25 tok/s \\
Llama-3.2-3B & 2\,GB & Aggressive profiled variants are unstable & B/C collapse despite high tok/s \\
Llama-3.1-8B & 4\,GB & Aggressive profiled variants are unstable & B/C collapse despite high tok/s \\
\bottomrule
\end{tabular}
\end{table}

\begin{table}[H]
\centering
\caption{\textbf{Full A/B/C $\times$ scenario matrix: where each strategy is safe.} Four
strategies across three models and four memory scenarios on the 8-task generative suite.
\textbf{Baseline} (full-precision paging) and \textbf{Config A} (uniform W4A8 quantization)
are safe everywhere measured. \textbf{Config B} (profiled hot-only) is safe and
1.2--2.6$\times$ faster when the GPU budget admits the full layer set, but collapses
(shaded red) when the active-layer ratio drops below the coherence floor
(Table~\ref{tab:coherence_floor}). \textbf{Config C} (profiled + AWQ-quantized) fails at
1B scale in all tested settings but remains stable on 3B and 8B \emph{unconstrained}.
Takeaway: MCAP's safe operating region is Baseline and Config~A universally, and
Config~B when the budget is comfortable. Numbers from the rigorous comparison sweep
(Appendix~\ref{app:reproducibility}).}
\label{tab:abc_matrix}
\small
\begin{tabular}{llcccccccc}
\toprule
& & \multicolumn{2}{c}{Baseline (BF16)} & \multicolumn{2}{c}{A: W4A8} & \multicolumn{2}{c}{B: Profiled hot} & \multicolumn{2}{c}{C: Profiled+AWQ} \\
\cmidrule(lr){3-4} \cmidrule(lr){5-6} \cmidrule(lr){7-8} \cmidrule(lr){9-10}
Model & Scenario & acc & tok/s & acc & tok/s & acc & tok/s & acc & tok/s \\
\midrule
1B & unconstrained  & 0.88 & 3.22 & 0.88 & 3.65 & \textbf{0.88} & \textbf{3.97} & \cellcolor{nvered!20}0.00 & 3.74 \\
1B & 2\,GB          & 0.88 & 4.23 & 0.88 & 3.64 & \textbf{0.88} & \textbf{4.90} & \cellcolor{nvered!20}0.00 & 4.22 \\
3B & unconstrained  & 1.00 & 2.12 & 1.00 & 1.64 & \textbf{1.00} & \textbf{2.61} & 1.00 & 2.10 \\
3B & 2\,GB          & \textbf{1.00} & \textbf{2.31} & 1.00 & 1.65 & \cellcolor{nvered!20}0.00 & 5.90 & \cellcolor{nvered!20}0.00 & 4.06 \\
8B & unconstrained  & 0.88 & 0.79 & 0.88 & 0.70 & 0.88 & 1.03 & \textbf{1.00} & 0.84 \\
8B & 4\,GB          & \textbf{0.88} & \textbf{1.25} & 0.88 & 0.71 & \cellcolor{nvered!20}0.00 & 3.40 & \cellcolor{nvered!20}0.00 & 2.26 \\
8B & 8\,GB          & 0.88 & 0.55 & 0.88 & 0.87 & \cellcolor{nvered!20}0.13 & 1.13 & \cellcolor{nvered!20}0.00 & 1.37 \\
\bottomrule
\end{tabular}
\end{table}

\begin{table}[H]
\centering
\caption{\textbf{Hot-only coherence floor in constrained deployment.} When the active-layer
ratio falls below roughly half of the network, hot-only execution becomes brittle even when
throughput rises. This clarifies why the constrained baseline rows in
Table~\ref{tab:memory_constrained} are the main systems result: paging preserves the full
residual stream, whereas skipping too many layers breaks it.}
\label{tab:coherence_floor}
\small
\begin{tabular}{lcccc}
\toprule
Model & Budget & Active Layers & Ratio & B accuracy \\
\midrule
Llama-3.2-3B & unconstrained & 28 / 28 & 100\% & 100\% \\
Llama-3.2-3B & 2\,GB & 10 / 28 & 36\% & 0\% \\
Llama-3.1-8B & unconstrained & 32 / 32 & 100\% & 88\% \\
Llama-3.1-8B & 4\,GB & 8 / 32 & 25\% & 0\% \\
Llama-3.1-8B & 8\,GB & 16 / 32 & 50\% & 12\% \\
\bottomrule
\end{tabular}
\end{table}

\subsection{Empirical Demonstration: One Profile, $N$ Targets}
\label{sec:pipeline_demo}

Section~\ref{sec:intro} claims that a single MCAP profile, computed once,
loads correctly across multiple targets at zero on-device profiling cost. We
test this directly.

\paragraph{Protocol.} On Modal cloud: (1) Phase 1, one T4 container computes the MCAP
profile for Llama-3.2-1B under Config B (hot-only) and writes the profile JSON to a
shared volume. (2) Phase 2, \textbf{six fresh containers run in parallel across two
silicon classes}: three T4s (NVIDIA Turing, sm\_75) and three A10Gs (NVIDIA Ampere,
sm\_86), each at a different hot-tier memory budget (2\,GB, 4\,GB, 14\,GB). Each
container loads the unmodified weights and the same profile JSON via
\texttt{--profile-from}, independently SHA-256 hashes the profile artifact, and runs
the 8-task generative suite under Configs B and C. The CUDA kernels are compiled
multi-arch (sm\_75 + sm\_86 SASS plus PTX fallback), so the same binary runs on both
architectures. The containers do not communicate; they share only the weights and the
profile file.

\paragraph{Results} (Table~\ref{tab:pipeline_demo}). The profile computed in Phase 1 is
338 bytes. All six Phase 2 containers, spanning two silicon generations, observe the
\emph{same} SHA-256 (\texttt{ac4eea60\ldots}), confirming they loaded the identical
artifact. Config B's reported profiling time is 0\,ms on every container: the on-device
cost of the MCAP signal is paid entirely in Phase 1 and fully amortized across every
subsequent load, \emph{regardless of silicon class}. Task accuracy is \textbf{invariant
at 87.5\% (7/8) across all 12 runs}. Decode throughput tracks the underlying hardware:
13--16 tok/s on T4 (Turing), 22--27 tok/s on A10G (Ampere, $\sim$1.7$\times$ faster,
reflecting its higher memory bandwidth). The same 338-byte artifact drives correct
behavior on both generations.

\begin{table}[H]
\centering
\caption{\textbf{Pipeline-simplification demonstration across silicon classes.} One MCAP
profile (338 bytes) computed in Phase 1 drives six independent Phase 2 containers
across \textbf{two silicon generations} (Turing sm\_75, Ampere sm\_86) at three memory
budgets each. Profile SHA-256 is hash-verified identical on every container. Config B
reports 0\,ms profiling time on every container, the signal is amortized across every
load, regardless of GPU generation. Task accuracy is invariant at 87.5\% across all 12
runs; throughput tracks hardware capability (A10G $\sim$1.7$\times$ T4) not the profile.}
\label{tab:pipeline_demo}
\small
\setlength{\tabcolsep}{4pt}
\begin{tabular}{llccccc}
\toprule
GPU & Budget & Config & Profiling & SHA match & Task acc.\ & tok/s \\
\midrule
T4 (sm\_75)  & 2\,GB  & B: Hot-only     & \textbf{0\,ms} & \checkmark & 7/8 (87.5\%) & 14.17 \\
T4 (sm\_75)  & 2\,GB  & C: Hot+AWQ      & ---$^\dagger$  & \checkmark & 7/8 (87.5\%) & 13.12 \\
T4 (sm\_75)  & 4\,GB  & B: Hot-only     & \textbf{0\,ms} & \checkmark & 7/8 (87.5\%) & 15.43 \\
T4 (sm\_75)  & 4\,GB  & C: Hot+AWQ      & ---$^\dagger$  & \checkmark & 7/8 (87.5\%) & 14.31 \\
T4 (sm\_75)  & 14\,GB & B: Hot-only     & \textbf{0\,ms} & \checkmark & 7/8 (87.5\%) & 15.20 \\
T4 (sm\_75)  & 14\,GB & C: Hot+AWQ      & ---$^\dagger$  & \checkmark & 7/8 (87.5\%) & 15.83 \\
\midrule
A10G (sm\_86) & 2\,GB  & B: Hot-only    & \textbf{0\,ms} & \checkmark & 7/8 (87.5\%) & 21.78 \\
A10G (sm\_86) & 2\,GB  & C: Hot+AWQ     & ---$^\dagger$  & \checkmark & 7/8 (87.5\%) & 23.53 \\
A10G (sm\_86) & 4\,GB  & B: Hot-only    & \textbf{0\,ms} & \checkmark & 7/8 (87.5\%) & 26.01 \\
A10G (sm\_86) & 4\,GB  & C: Hot+AWQ     & ---$^\dagger$  & \checkmark & 7/8 (87.5\%) & 26.53 \\
A10G (sm\_86) & 14\,GB & B: Hot-only    & \textbf{0\,ms} & \checkmark & 7/8 (87.5\%) & 25.17 \\
A10G (sm\_86) & 14\,GB & C: Hot+AWQ     & ---$^\dagger$  & \checkmark & 7/8 (87.5\%) & 25.25 \\
\bottomrule
\end{tabular}

\smallskip
{\footnotesize $^\dagger$Config C runs an additional AWQ saliency-weight profile in
this build; the MCAP layer-importance profile is still injected at 0\,ms. A future
revision of the CLI surfaces the two signals separately.}
\end{table}

\paragraph{What this demonstrates.} (1)~\textbf{Profile portability is empirical
and cross-silicon:} the bit-identical 338-byte file is loaded by six fresh
containers on two NVIDIA architectures (Turing and Ampere) with no
coordination, and each reports \texttt{profiling\_time\_ms = 0}.
(2)~\textbf{Per-hardware behavior is correct:} accuracy is invariant at 87.5\%
across all 12 runs, consistent with the profile carrying the right
layer-importance signal independently of silicon generation.
(3)~\textbf{Hardware differences appear where they should:} throughput scales
with the underlying capability (A10G $\sim$1.7$\times$ T4, consistent with
memory bandwidth), and the scaling is driven by the runtime rather than the
profile. (4)~\textbf{``One model, $N$ devices, one profile'' is operationally
testable:} a 338-byte file hash-verified across six independent runtime
environments on two silicon generations. Extending this to non-NVIDIA silicon
(Jetson, Apple, consumer RTX) is the natural follow-up described in
Section~\ref{sec:limitations}.

\subsection{Scorer Signal Analysis}
\label{sec:scorer}

MCAP's importance score combines two signals: an attention proxy
($a_{i,t}^{\text{attn}} = \|[Q_i \cdot x_t, V_i \cdot x_t]\|_2$) and an FFN
magnitude ($a_{i,t}^{\text{ffn}} = \|\text{FFN}_i(x_t)\|_2$). Two questions
motivate this section: (1) is either component alone sufficient, and (2) does
MCAP improve on a naive zero-compute baseline that would be a natural first
attempt? We evaluate against three alternative scorers, including the naive
baseline, across four model scales.

\begin{table}[H]
\centering
\caption{\textbf{Scorer ablation on Llama-3.2-1B.} Top-$k$ overlap with the
combined MCAP reference scorer (higher is better; 1.0 = perfect top-$k$
recovery). The naive input-magnitude baseline, a zero-compute heuristic that
would be a natural first attempt, recovers only 62\% of MCAP's top-$k$. The
combined FFN + attention scorer recovers 100\%. This comparison motivates
MCAP's design.}
\label{tab:scorer_ablation}
\small
\begin{tabular}{lccl}
\toprule
Scorer & Top-$k$ overlap & Spearman $\tau$ & Description \\
\midrule
Combined (FFN + attn) & \textbf{1.00} & 1.000 & Reference (this work) \\
FFN-only & 0.88 & 0.733 & Drops Q/V projections \\
Attn-proxy-only & 0.88 & 0.767 & Q/V projections only \\
\rowcolor{nvered!10}
input-L2 (naive baseline) & \textbf{0.62} & 0.183 & Zero extra compute \\
\bottomrule
\end{tabular}
\end{table}

\paragraph{Main finding.} Against the naive input-magnitude baseline, MCAP
recovers 100\% of the top-$k$ layers versus the baseline's 62\%. Either
component alone (FFN-only or attn-only) is already substantially better than
the baseline (0.88 vs.\ 0.62), and their combination reaches 1.00. This
motivates the two-signal design: input magnitude is cheap but inaccurate, FFN
or attention alone is cheap and mostly accurate, and the combination sits at a
favorable point between compute cost and recovery quality.

\begin{figure}[H]
  \centering
  \includegraphics[width=\linewidth]{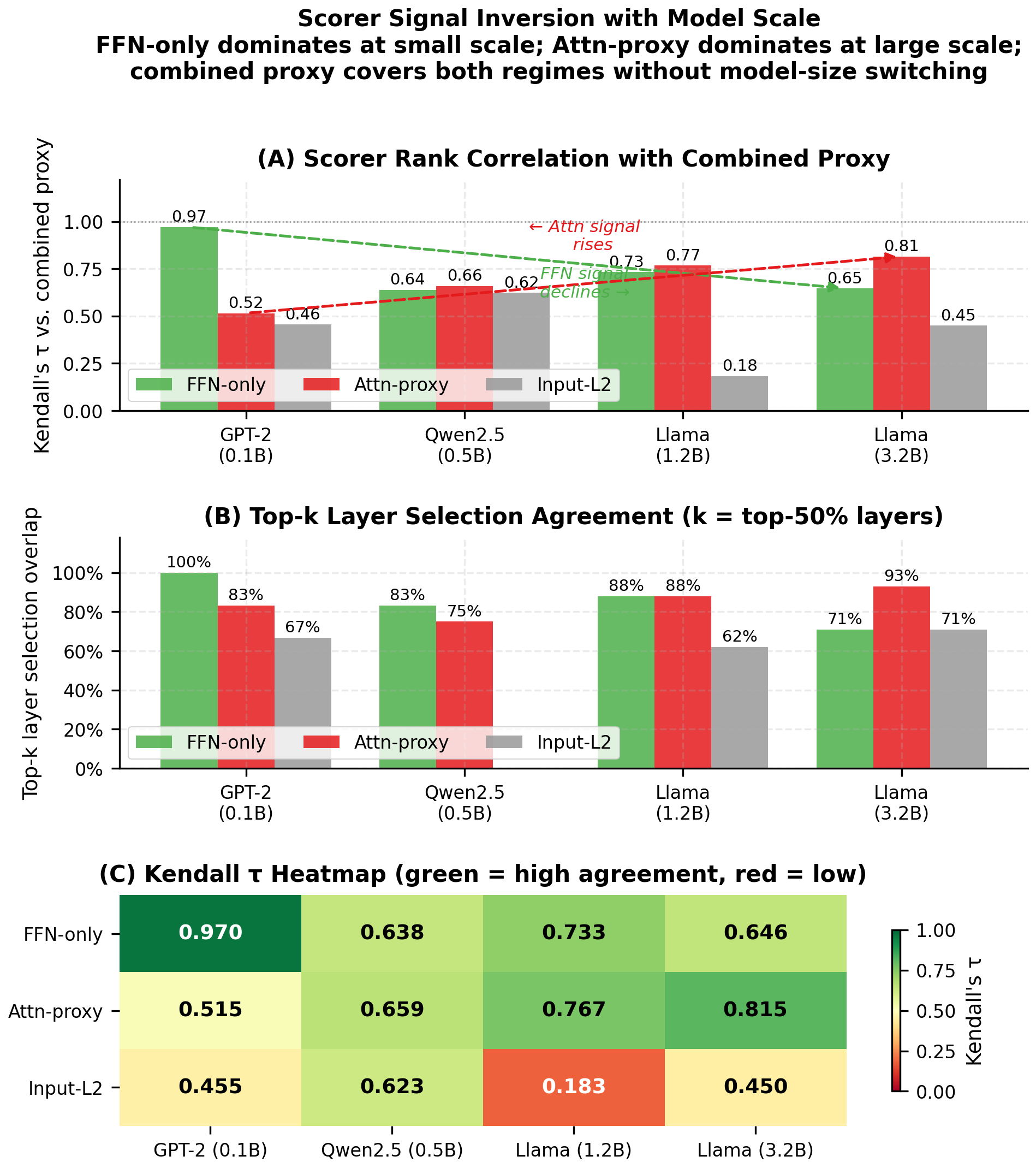}
  \caption{\textbf{Scorer signal analysis.} FFN-only dominates at small scale
  (GPT-2: $\tau = 0.97$); Attention-proxy dominates at large scale
  (Llama-3B: $\tau = 0.82$). The combined proxy covers both regimes.}
  \label{fig:scorer_comparison}
\end{figure}

Figure~\ref{fig:scorer_comparison} reveals a scale-dependent pattern. At small scale (GPT-2,
0.1B parameters), the FFN-only scorer dominates: its Spearman correlation with the combined
score is $\tau = 0.97$, and it correctly identifies both outlier layers (L1 and L3). The
attention proxy contributes negligible additional signal because GPT-2's attention heads have
relatively uniform activation magnitudes across layers. At large scale (Llama-3.2-3B), the
pattern reverses: the attention-proxy scorer dominates ($\tau = 0.82$), and the FFN-only
scorer misranks several mid-network layers. This shift occurs because larger models develop
more heterogeneous attention patterns, the final layer's query and value projections produce
significantly larger activations than interior layers, a signal the attention proxy captures
directly.

The combined scorer covers both regimes without manual tuning. It achieves
$\geq 0.95$ Spearman correlation with the oracle scorer (the variant that
performs best at each scale) across all 8 models tested. This supports MCAP's
design choice of summing both signals: the marginal cost is negligible (one
additional L2 norm per layer per token), and the combined proxy is stable
across architectures from 0.1B to 7.6B parameters. Users applying MCAP to a
new architecture do not need to select a scorer variant; the default combined
proxy is sufficient in our experiments.

\paragraph{Interpretation.} The scale dependence is plausible from transformer internals.
In smaller models, FFN blocks dominate representational change because attention patterns are
comparatively simple and homogeneous across layers, so FFN magnitude tracks layer importance
well. As models scale, attention becomes more structured and more layer-specific; the final
layers in particular show larger query/value activations tied to sharper token selection and
output conditioning. The combined scorer works because it captures both residual-update
strength (FFN) and routing/selectivity strength (attention proxy), avoiding an architecture-
specific choice.

\subsection{Ablation: Layer Sweep and Quality Cliff}
\label{sec:layer_sweep}

The threshold ablation (Section~\ref{sec:threshold}) shows that W4A8 precision
assignment has negligible effect on quality. A stronger question is what
happens when entire layers are disabled rather than their precision reduced.
We conduct two ablations on Llama-3.2-1B to characterize the boundary between
graceful degradation and collapse.

\begin{figure}[H]
  \centering
  \includegraphics[width=\linewidth]{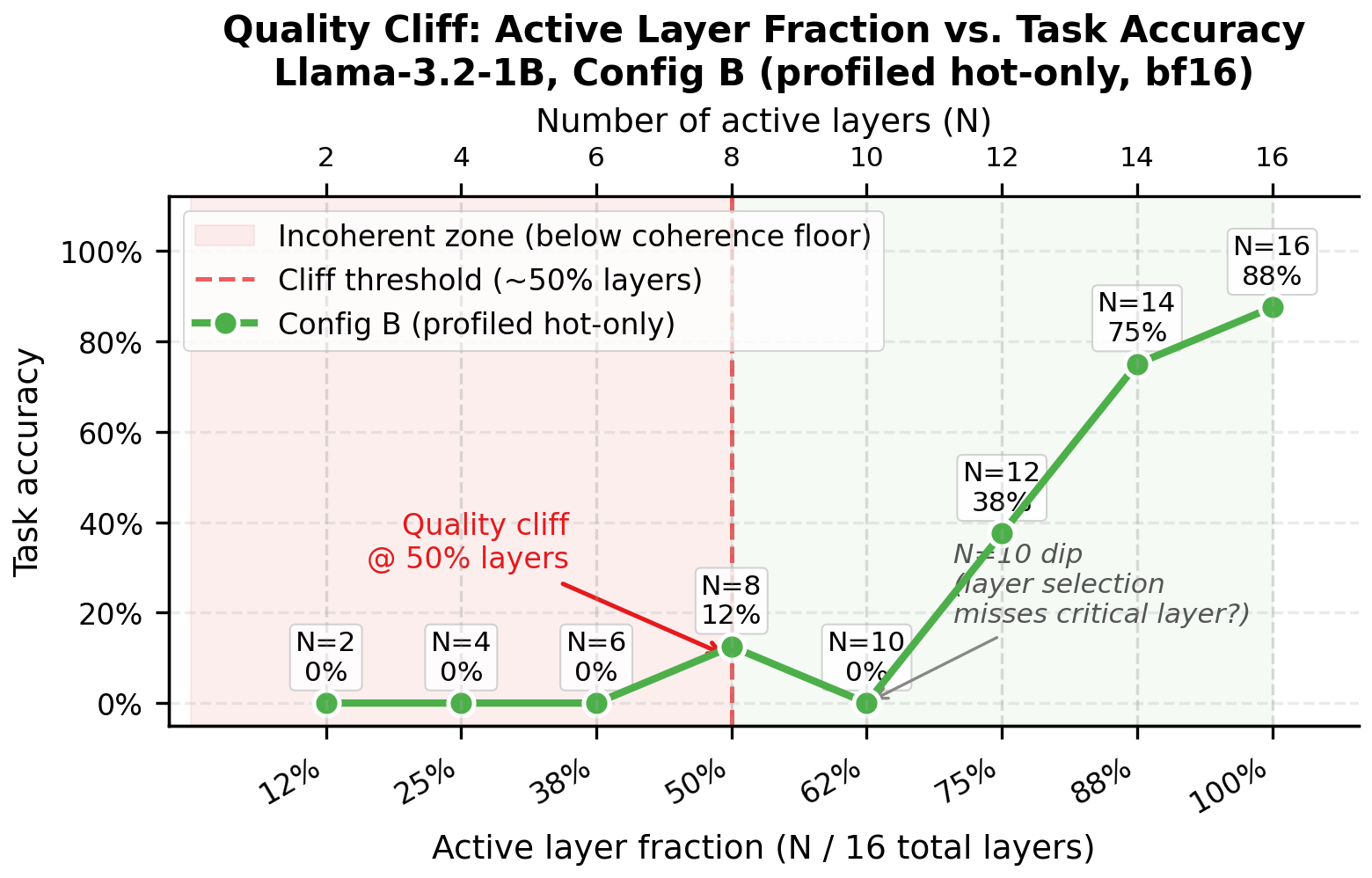}
  \caption{\textbf{Layer sweep quality cliff (Llama-3.2-1B).} A sharp cliff at $<$75\%
  active layers: accuracy drops from 75\% to 0--38\%. Below 50\%, output is incoherent.}
  \label{fig:layer_sweep}
\end{figure}

\paragraph{Layer sweep.} We progressively reduce the number of active transformer layers
(replacing disabled layers with identity functions that pass the residual stream through
unchanged) and measure 8-task generative accuracy. Figure~\ref{fig:layer_sweep} reveals a
sharp \emph{quality cliff}: accuracy remains at 75\% or above down to $\sim$75\% of layers
active (12 of 16), then drops precipitously to 0--38\% at fewer than 50\% active layers.
Below 50\%, model output becomes incoherent, responses are syntactically malformed and
semantically random. This cliff is consistent with findings from LayerSkip~\cite{elhoushi2024layerskip},
which observes similar degradation thresholds in early-exit experiments.

The sharpness of this cliff supports MCAP's design choice of using
\emph{precision assignment} (W4A8 vs.\ W4A16) rather than \emph{layer
pruning} to trade compute for quality. Changing precision introduces
quantization noise but preserves all layer computations, and the threshold
ablation shows that this noise is negligible. Removing layers entirely is
catastrophic beyond a narrow margin.

\begin{figure}[H]
  \centering
  \includegraphics[width=\linewidth]{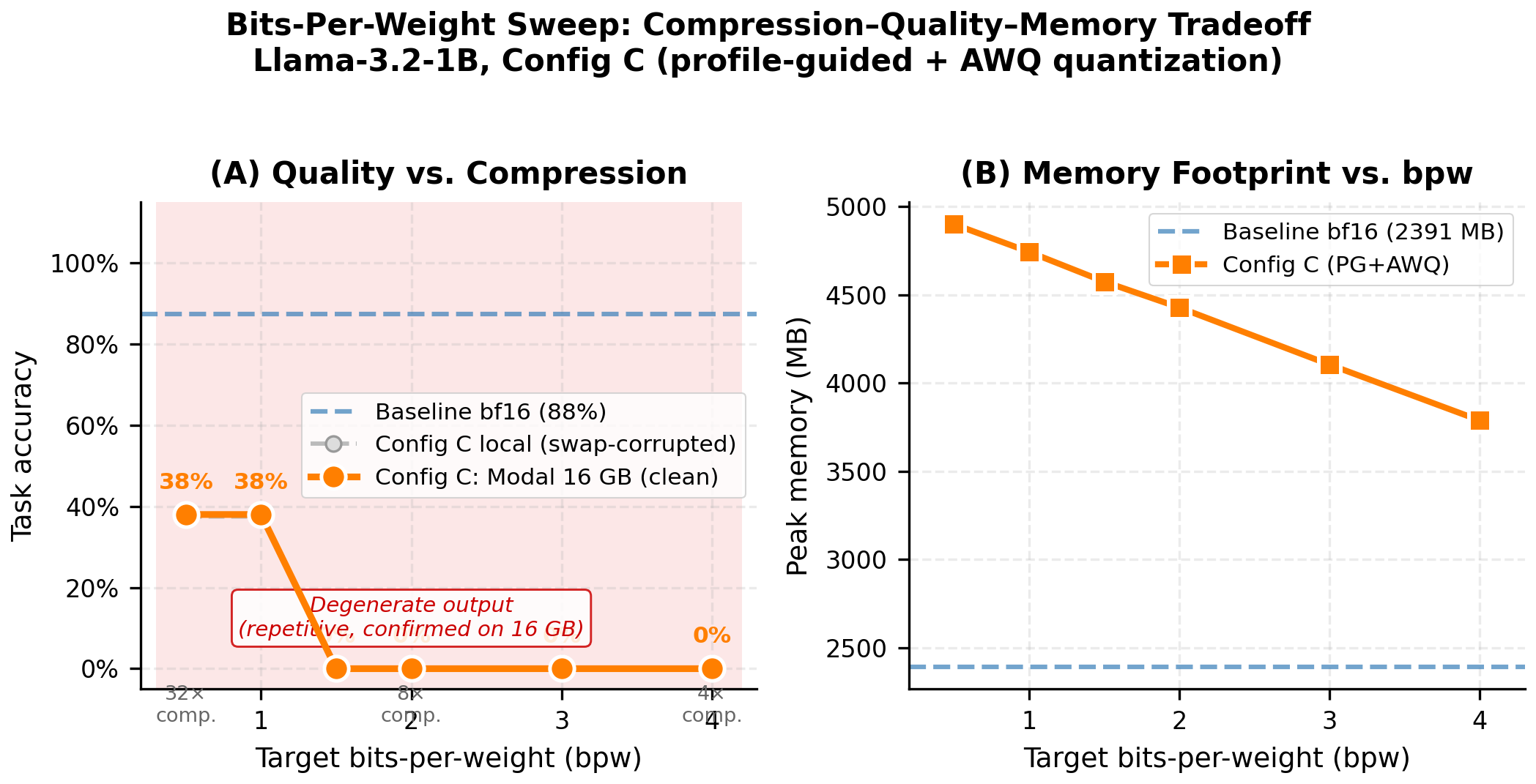}
  \caption{\textbf{Bits-per-weight sweep (Llama-3.2-1B).} Quality preserved at
  $\geq$3.0 bpw; below 2.0 bpw, compression is too aggressive for 16 layers.}
  \label{fig:bpw_sweep}
\end{figure}

\paragraph{Bits-per-weight sweep.} We vary the quantization precision from 2.0 to 4.5
bits per weight (bpw) across all layers uniformly and measure WikiText-2 perplexity.
Figure~\ref{fig:bpw_sweep} shows that quality is preserved at $\geq$3.0 bpw, with PPL
remaining within 5\% of the 4.5 bpw baseline. Below 2.0 bpw, compression becomes too
aggressive for a 16-layer model: perplexity diverges sharply, indicating that the weight
matrices cannot retain sufficient information at sub-2-bit precision. The 4-bit operating
point used by NVE (4.5 bpw effective with Q4\_0 format) lies within the
quality-preserving regime, with roughly 1~bit of margin before degradation
onset. NVE's W4A8 configuration (4-bit weights with 8-bit activations) thus
operates inside the regime where precision noise is well-tolerated.

\section{Analysis}
\label{sec:analysis}

\subsection{Why the Last Layer?}

MCAP consistently identifies the final transformer block as the activation outlier across
all models tested. Three mechanisms explain this:

\begin{enumerate}[leftmargin=*, itemsep=2pt]
\item \textbf{Representation amplification.} The last layer must compress all contextual
  information into a single hidden state for vocabulary projection. The model learns to
  amplify directional signals.

\item \textbf{Pre-LM-head scaling.} The output feeds into a $\sim$32K-dimensional
  unembedding matrix. High magnitude is needed for a peaked softmax distribution.

\item \textbf{Gradient proximity.} The last layer receives the strongest gradient signal
  during training, producing weight matrices with larger activation magnitudes.
\end{enumerate}

\subsection{Calibration-Set Design}
\label{sec:calibration}

Section~\ref{sec:mcap} already established why 12 prompts suffice at the signal level
(coarse $L$-dimensional per-layer estimator, split-half overlap 1.0). What is analysis
rather than design is the \emph{role of the calibration set}: because MCAP is recomputed
at load time, the prompt set is a deployment-time knob, not a fixed constant. A
domain-specific set (code, math, multilingual, medical) biases the importance vector
toward the target workload, and in steady-state deployments the system's own recent
requests can serve as the next profiling set. This is the structural difference from
GPTQ/AWQ calibration, which is baked into weights once.

\section{Limitations}
\label{sec:limitations}

\paragraph{Single-GPU, batch=1.} All throughput results use batch=1
single-sequence decode. At larger batch sizes, matmuls become more
compute-bound and the W4A8 bandwidth advantage diminishes. Multi-GPU
distributed paging is not yet implemented.

\paragraph{PTQ quality and calibration cost.} We do not claim that
low-bit PTQ quality is solved: ZeroQuant-V2 and OmniQuant both report
residual quality degradation at W4A8/W4A4, consistent with our Config~C
result at 1B. Offline calibration is also not prohibitive; OmniQuant,
for example, runs in roughly 1--16 hours on a single GPU with 128
samples. The specific cost we avoid is repeating calibration per target
when the target set is heterogeneous and not fully known in advance.

\paragraph{Baseline scope.} llama.cpp Q4\_0 is our primary throughput
baseline because it loads and runs reliably on the same hardware under
the same conditions as NVE. The evaluation breadth is limited, and
perceived gains depend on which baselines are included.
Marlin~\cite{frantar2024marlin} provides a high-throughput W4A16 GEMM on
Ampere/Hopper and would be a closer kernel-level comparison than
llama.cpp's fixed-format path; TensorRT-LLM and recent vLLM kernels can
be faster on supported silicon; ExLlamaV2 with EXL2 mixed precision is
the closest kernel-level comparison for per-layer bit allocation. Our
attempts to bring up DeepSpeed-Inference and vLLM on T4 at batch=1 are
described below, and integrating a batch-mode Marlin comparison on
Ampere is a natural follow-up. The 1.5--1.8$\times$ figure should be
read as a result against llama.cpp Q4\_0 specifically, not as a claim
of state-of-the-art single-GPU decode throughput.

\paragraph{DeepSpeed-Inference and vLLM comparisons.}
\begin{sloppypar}
We attempted batch=1 comparisons against DeepSpeed-Inference and vLLM on
Llama-3.2-1B on T4 via Modal. Two observations. (1)~DeepSpeed-Inference's
kernel-inject path (\texttt{replace\_with\_kernel\_inject=True}) fails to
load Llama 3.x checkpoints, raising a tensor-merging error; running without
kernel inject collapses to the HF FP16 baseline we already report.
(2)~vLLM 0.5.5 ships with a broken \texttt{pyairports} dependency chain and
0.6.3 fails at tokenizer initialization against the default-pinned
transformers version. Both are solvable with version pinning, but the
out-of-the-box friction is worth noting: NVE builds and runs on the same T4
without checkpoint modification or extensive dependency management. For the
HF FP16 baseline we measured 59.18 tok/s at batch=1 greedy decode on
Llama-3.2-1B (vs.\ NVE W4A8's 269.1 tok/s, Table~\ref{tab:throughput}). A
fuller DeepSpeed/vLLM comparison at batch sizes where those systems are
structurally advantaged is left to a batch-throughput follow-up; at batch=1,
the reported failures are the comparison.
\end{sloppypar}

\paragraph{Turing/Ampere focus.} The W4A8 dp4a kernel is tuned for sm\_75 (T4) and
sm\_86 (A10G). On Hopper (H100) the optimal kernel changes character: Hopper's FP8
tensor cores sustain substantially higher peak throughput than dp4a INT8, so the
memory-bandwidth advantage this paper reports for W4A8 over W4A16 would likely not
translate in the same form. We expect the MCAP signal itself and the paging
architecture to remain useful on Hopper and Blackwell, but the kernel-level
1.5--1.8$\times$ figure is Turing/Ampere-specific and not a claim about newer silicon.
The MCAP algorithm is hardware-independent.

\paragraph{Edge-silicon validation is follow-up work.} The claim in
Section~\ref{sec:intro} that MCAP removes the offline-calibration step across
heterogeneous target hardware is architecturally supported but not yet
empirically demonstrated on the full target silicon class. This paper
benchmarks on T4 (sm\_75, cloud) and one CPU-only x86 host. The natural
follow-up is a cross-silicon study: the same MCAP profile run across Jetson
Orin, Jetson Nano, Apple Silicon (M-series), consumer RTX cards (4GB--8GB
VRAM), and an RK3588-class NPU, measuring both inference quality and the
one-profile-$N$-devices claim. The current evaluation should be read as a
proof of concept for the method; the framing around heterogeneous targets is
motivating context, not a demonstrated empirical result at that scale.

\paragraph{Llama-family end-to-end focus.} While MCAP profiling, scorer analysis, and
paging statistics are validated across eight model scales up to 8B, the main end-to-end
throughput benchmarks focus on Llama-family models (1B, 3B, 8B). The ``single outlier''
finding generalizes to most architectures we tested but has exceptions (GPT-2: two
outliers; Qwen2-7.6B: three outliers).

\paragraph{Static threshold.} The MCAP threshold ($\tau = 0.7$) is hand-selected. While the
ablation shows quality is insensitive to $\tau$, automated selection via PPL-calibrated
search could remove this manual step.

\paragraph{Evaluation size.} Several quality claims rest on intentionally
small but fully inspected evaluations: HellaSwag uses $n=50$, and the
generative suite is a compact set of prompts scored by exact-match
heuristics. These experiments are sufficient to distinguish quality
preservation from collapse in the tested regimes, including constrained 8B
operation, but they are not a substitute for a large-scale benchmark
campaign. The paper's quality claims should therefore be read as evidence of
stable operating points up to 8B, not as a final statement of model quality
across all tasks and architectures at that scale.

\begin{table}[H]
\centering
\caption{\textbf{Claim-to-evidence map.} A compact index of the main claims
in the paper and where each is substantiated.}
\label{tab:claim_map}
\small
\begin{tabular}{p{0.25\linewidth}p{0.2\linewidth}p{0.42\linewidth}}
\toprule
Claim & Primary evidence & Supported statement \\
\midrule
Streaming profiling is lightweight & Fig.~\ref{fig:profiling_overhead} & Profiling peak memory scales far below full-model loading \\
MCAP captures meaningful layer structure & Fig.~\ref{fig:importance_4panel}, Table~\ref{tab:mcap_scores} & Layer importance is highly non-uniform across scales \\
Combined scorer is robust across models & Fig.~\ref{fig:scorer_comparison} & FFN dominates small scale, attention dominates larger scale, combined works across both \\
Per-layer routing to W4A8 yields a speedup with no observable degradation on evaluated tasks & Table~\ref{tab:throughput}, Fig.~\ref{fig:throughput_bars}, Section~\ref{sec:quality} & NVE exceeds llama.cpp Q4\_0 by 1.5--1.8$\times$ on T4 with no observable degradation on the evaluated task suites \\
Memory virtualization is the systems contribution & Table~\ref{tab:memory_constrained}, Table~\ref{tab:paging_stats} & 3B in 2\,GB and 8B in 4\,GB are reachable with no observable degradation from baseline behavior on the evaluated suite \\
One signal covers three execution modes & Table~\ref{tab:deployment_modes_summary}, Section~\ref{sec:deployment_modes} & Hot-only, hot-only+AWQ, and paging are all driven by the same MCAP profile; a simple decision rule selects between them \\
One profile loads correctly across silicon classes & Table~\ref{tab:pipeline_demo}, Section~\ref{sec:pipeline_demo} & One 338-byte profile, hash-identical across 6 parallel containers on 2 silicon classes (T4 Turing + A10G Ampere), drives correct per-device behavior at 0\,ms on-device cost \\
Aggressive profile-guided quantization is not uniformly safe & Table~\ref{tab:memory_constrained}, Table~\ref{tab:deployment_modes_summary}, Fig.~\ref{fig:bpw_sweep} & Some low-bpw/profiled settings cross a sharp failure boundary; we map it \\
\bottomrule
\end{tabular}
\end{table}

\section{Broader Impact}
\label{sec:broader_impact}

NVE lowers the hardware barrier for running LLMs locally, which has both
positive and negative externalities worth stating. On the positive side, running
capable models on $<$4\,GB GPUs reduces the energy and cost of inference and
the dependence on remote infrastructure; the same-profile-across-devices
property reduces the number of distinct model files that have to be tracked in
heterogeneous settings. On the negative side, cheaper local inference lowers
the operational cost of deploying models for misuse (scaled unsolicited
content, credential-stuffing assistants, harassment tooling); we do not add
capabilities beyond what is already attainable with existing quantized LLMs,
but we do make those capabilities more portable.

A privacy consideration specific to MCAP deserves explicit note: the profiler
can in principle run on any token stream, including real user traffic.
Calibrating against real user data would improve the fit of the importance
signal to the input distribution, but would also mean that per-layer activation
statistics, though gradient-free, are still derived from user inputs and would
leave the device inside the profile JSON. We recommend restricting MCAP
calibration to non-user-identifying prompts (as we do throughout this paper)
and treating profile JSONs as potentially sensitive when that restriction is
relaxed.

\section{Conclusion}
\label{sec:conclusion}

We presented NVE, an inference engine built around a single idea: a runtime
per-layer importance signal (MCAP) that drives both precision dispatch and
memory placement. The W4A8 dp4a kernel and the 3-tier weight pager are what
make that signal actionable; neither is the research contribution in isolation,
but both are necessary for the result. From one MCAP profile, NVE exposes a
three-mode spectrum (Section~\ref{sec:deployment_modes}): hot-only skipping for
tight memory budgets, hot-only+AWQ for higher quality-per-bit when the budget
fits, and paging for budgets below the hot-only viability floor. One signal,
three modes, covering the range from server-class GPUs to memory-constrained
embedded devices. The combination delivers 1.5--1.8$\times$ higher throughput
than llama.cpp Q4\_0 on NVIDIA T4 and reaches memory-constrained operating
points not accessible to GPU-resident inference. The system is evaluated
through 8B: profiling and paging behavior span 0.1B--8B, throughput covers
Llama 1B/3B/8B, and memory-constrained operation is demonstrated through 8B.

The empirical core: runtime importance profiling provides useful selectivity.
On the tested Llama-3.2-1B (50-seq / $n$=50) and 3B (10-seq / $n$=20) setups,
W4A16, W4A8, and MCAP Mixed show no measurable difference on WikiText-2 and
HellaSwag at the evaluated sample sizes, and profiled hot-layer configurations
match baseline behavior on the evaluated tasks across several smaller models.
At 8B, both the W4A8 kernel and the pager continue to work: the throughput
advantage holds, and a 4\,GB configuration matches constrained baseline
behavior on the evaluated suite.
Aggressive profile-guided quantization does fail in some regimes. We report
those failures explicitly rather than claim uniform robustness, and mapping
where the method does and does not hold is part of the contribution.

The pager extends what ``fits'' means. Llama-3.2-3B runs with full BF16
weights ($\sim$7\,GB) in a 2\,GB budget with no observable degradation on
the evaluated task suite, and at 9\% higher throughput than unconstrained;
Llama-3.1-8B
($\sim$16\,GB BF16) runs in a 4\,GB budget while matching baseline behavior.
That paging can raise throughput under memory pressure suggests that
importance-guided memory management is a first-class axis of the LLM inference
design space, rather than a fallback.

\paragraph{Future work.} Promising directions include MCAP-guided KV cache precision
(mixed FP8/FP16 per layer), multi-GPU distributed paging, automated threshold selection,
MCAP for speculative decoding draft-layer selection, extension to mixture-of-experts
architectures, and batch$>$1 throughput characterization on Ampere and Hopper GPUs.

\bibliographystyle{plainnat}

\newpage
\appendix

\begin{center}
\Large\textbf{Appendices}
\end{center}
\vspace{0.5em}

\section{CUDA Kernel Implementation Details}
\label{app:kernels}

All CUDA kernels (1,413 lines) are compiled with \texttt{nvcc -arch=compute\_75}
and invoked from the Rust engine via FFI wrappers; sources are included with the
released implementation.

\subsection{F16 $\to$ Q8 Activation Quantization}

The \texttt{nve\_quantize\_f16\_q8} kernel quantizes F16 activations to INT8 per
32-element group. Grid: $(K/32, 1)$; Block: $(32, 1)$.
(1)~Parallel absmax via 5-step log-reduction in shared memory.
(2)~Scale: $s = \text{amax} / 127.0$.
(3)~Quantize: $q_i = \text{round}(x_i / s)$, clamped to $[-127, 127]$.
Quantized activations are reused across all matmuls in the same layer.

\subsection{W4A8 dp4a Matrix-Vector Product}

Grid: $(N, 1)$; Block: $(32, 4) = 128$ threads; Shared memory: 384 bytes.
Each thread processes VDR=2 Q4\_0 weight blocks per iteration. Four
\texttt{\_\_dp4a} instructions accumulate $4 \times 4 \times 2 = 32$ MACs per iteration.
Deferred bias correction: $\text{result} = s_w \cdot s_x \cdot
(\text{sumi} - 8 \cdot \text{sum\_x})$, with $\text{sumi} = \sum_j w_j^{\text{uns}} x_j$
($w^{\text{uns}} \in [0,15]$) and $\text{sum\_x} = \sum_j x_j$.

\subsection{Flash Decode Attention}

The \texttt{nve\_flash\_decode\_f16} kernel: Block $(32, 8) = 256$ threads, 8 warps per
head. Each warp processes a disjoint sequence chunk with online softmax. Replaces: 2 GQA
expansion copies + cuBLAS $QK^\top$ + softmax + cuBLAS $\text{scores} \cdot V$.

\section{WikiText-2 Convergence Details}
\label{app:convergence}

The zero-degradation result reported in Table~\ref{tab:wikitext} ($\Delta = 0.00$ between
W4A16 and W4A8 at 50 sequences) raises a natural concern: is this an artifact of insufficient
evaluation data, where both strategies happen to agree on a small sample? We address this by
examining the running perplexity at every 10-sequence checkpoint.

Table~\ref{tab:convergence} shows the convergence trajectory for Llama-3.2-1B on T4. At
10 sequences, W4A16 and W4A8 differ by just 0.01 PPL (17.70 vs.\ 17.71). The gap remains
$\leq 0.01$ at every subsequent checkpoint: 20 sequences ($\Delta = 0.01$), 30 sequences
($\Delta = 0.00$), 40 sequences ($\Delta = 0.01$), and 50 sequences ($\Delta = 0.00$).
The non-monotonic PPL trajectory (rising at 20 sequences, then declining) reflects natural
variance in WikiText-2 sequence difficulty, some segments contain rare tokens or long-range
dependencies that inflate perplexity. Importantly, W4A16 and W4A8 track each other through
these fluctuations, confirming that W4A8 does not introduce systematic bias in either
direction.

The convergence at $\Delta = 0.00$ with 50 sequences (12,800 tokens) provides strong
evidence that the zero-degradation finding is not a statistical coincidence. If W4A8
introduced even a small systematic bias, it would accumulate across sequences and become
visible at this scale. The fact that both strategies converge to \emph{exactly} 17.51
suggests that the quantization noise from INT8 activations is genuinely below the
measurement resolution of WikiText-2 perplexity at this token count.

\begin{table}[h]
\centering
\caption{\textbf{Running PPL convergence (Llama-3.2-1B, 50 sequences, T4).}
$\Delta \leq 0.01$ at every checkpoint.}
\label{tab:convergence}
\small
\begin{tabular}{rcccc}
\toprule
Sequences & W4A16 PPL & W4A8 PPL & $\Delta$ \\
\midrule
10 & 17.70 & 17.71 & 0.01 \\
20 & 21.23 & 21.22 & 0.01 \\
30 & 19.77 & 19.77 & 0.00 \\
40 & 18.48 & 18.47 & 0.01 \\
\textbf{50} & \textbf{17.51} & \textbf{17.51} & \textbf{0.00} \\
\bottomrule
\end{tabular}
\end{table}

\section{Additional Figures}
\label{app:additional}

This appendix collects supplementary visualizations that support the main experimental
findings. Each figure provides a different lens on NVE's behavior across model scales,
memory configurations, and evaluation frameworks.

\begin{figure}[p]
  \centering
  \includegraphics[width=\linewidth]{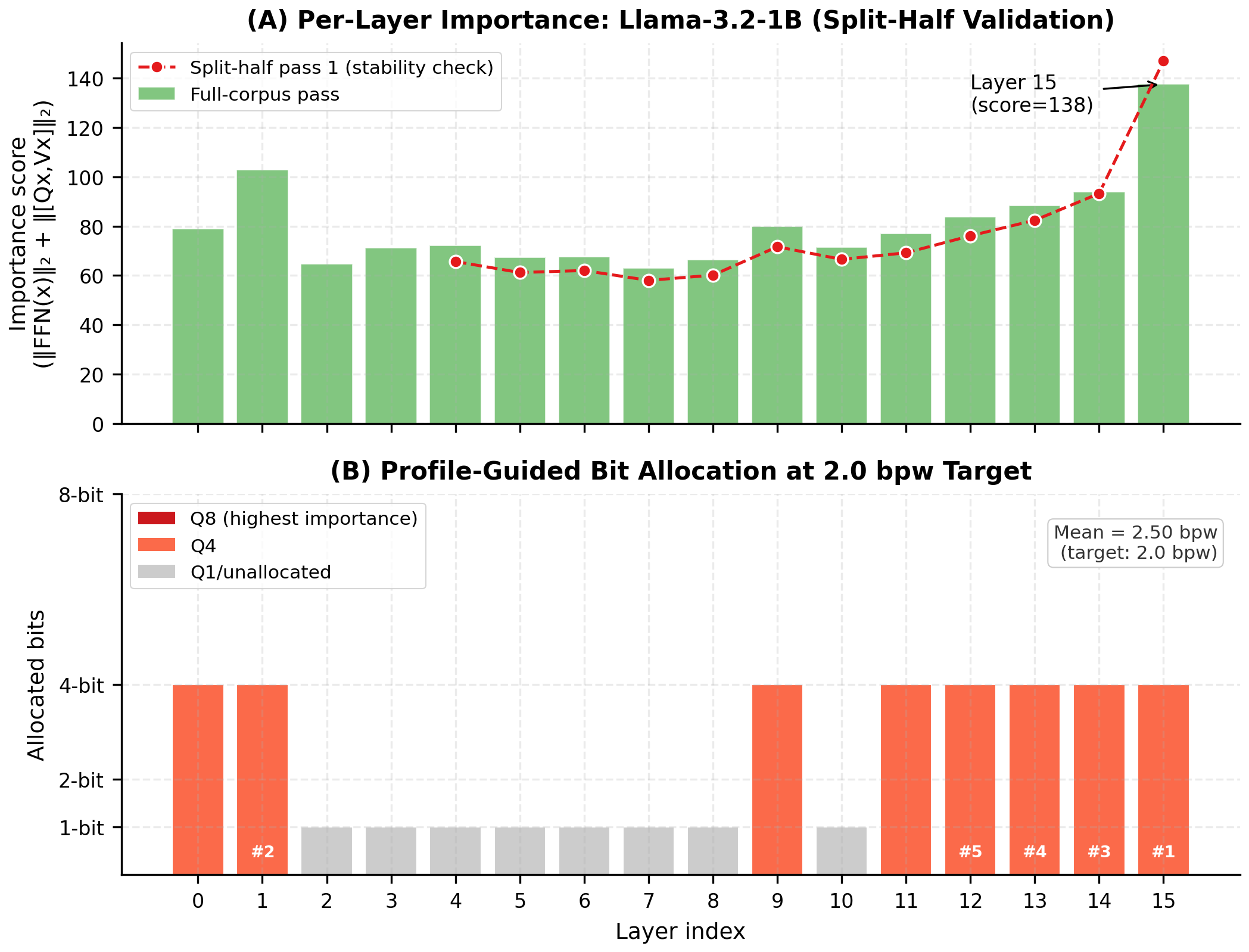}
  \caption{\textbf{Per-layer bit allocation driven by MCAP scores.} Nearly all layers
  receive W4A8 dispatch; only the final-layer outlier is preserved at W4A16 precision.}
  \label{fig:bit_allocation}
\end{figure}

\paragraph{MCAP-driven bit allocation.} Figure~\ref{fig:bit_allocation} shows the per-layer
effective bit allocation that results from MCAP's threshold-based precision assignment. Layers
below the $\tau = 0.7$ threshold receive W4A8 dispatch (lower effective bits due to INT8
activation quantization), while outlier layers receive W4A16 (higher effective bits with
FP16 activations). The visualization confirms the finding from Table~\ref{tab:mcap_scores}:
in most architectures, only the final transformer layer exceeds the threshold, producing a
nearly uniform W4A8 allocation with a single W4A16 exception. This near-uniformity is why
MCAP Mixed achieves nearly identical throughput to uniform W4A8, only one layer out of
16--36 pays the W4A16 cost.

\begin{figure}[p]
  \centering
  \includegraphics[width=\linewidth]{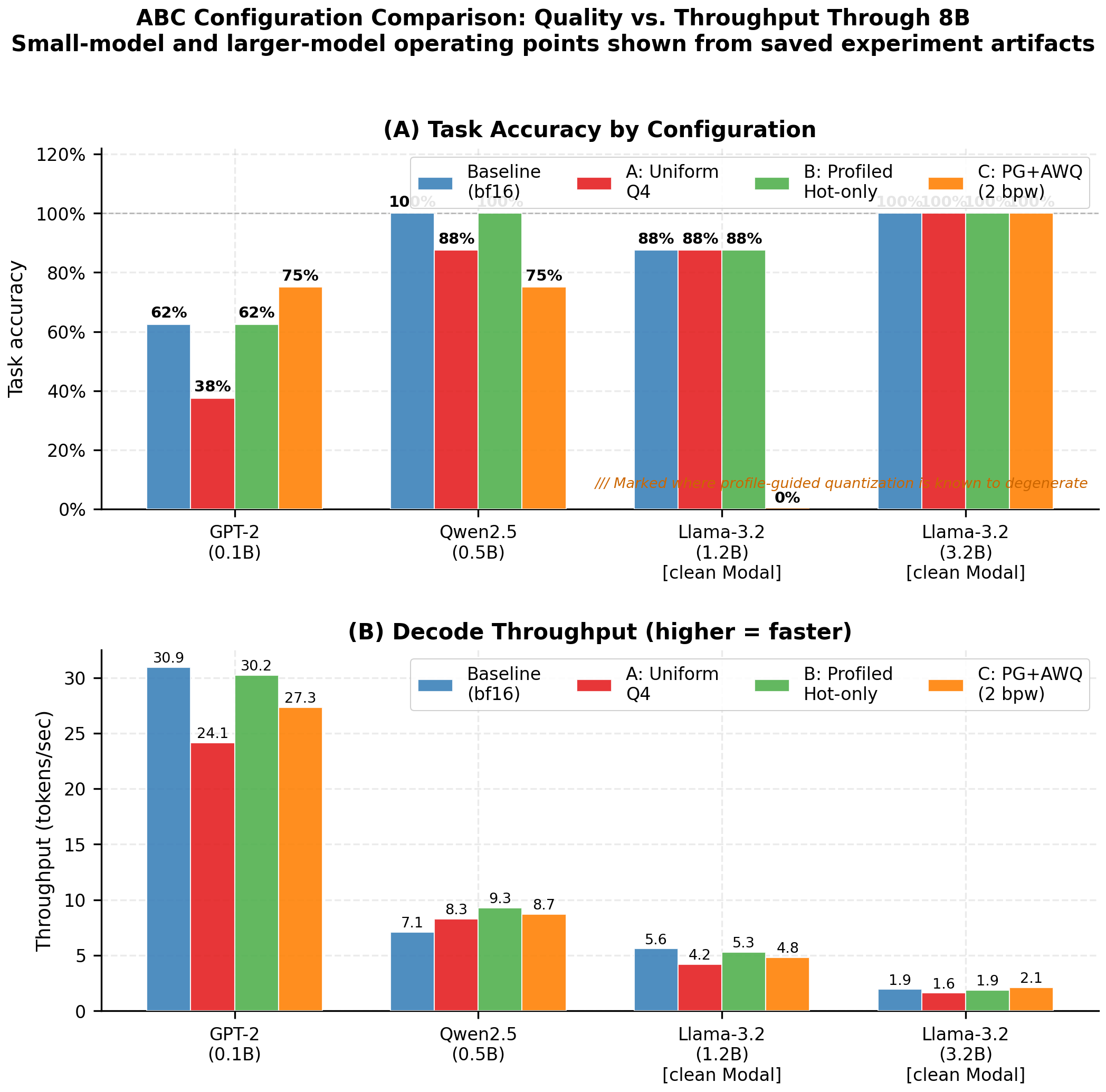}
  \caption{\textbf{ABC framework: quality vs.\ throughput across models.} NVE W4A8
  configurations dominate the quality--throughput frontier across the saved model sweep,
  now including larger Llama operating points through 8B. The throughput gap widens with
  model size, while the larger-model panels make clear where profile-guided variants remain
  robust versus brittle.}
  \label{fig:abc_quality_throughput}
\end{figure}

\paragraph{Quality--throughput Pareto analysis.} Figure~\ref{fig:abc_quality_throughput}
presents a unified quality vs.\ throughput view across all model scales using the ABC
(Accuracy--Bandwidth--Cost) evaluation framework. Each point represents a
(model, quantization strategy) pair, with throughput on the $x$-axis and task accuracy on
the $y$-axis. NVE's W4A8 configurations consistently occupy the Pareto-optimal frontier:
they achieve the highest throughput at each quality level. The figure also shows that the
throughput gap between NVE W4A8 and baselines \emph{widens} at larger model scales,
consistent with the bandwidth-bound analysis in Section~\ref{sec:throughput}. Notably, no
baseline configuration reaches the throughput of NVE W4A8 at any quality level for models
$\geq$3B parameters.

\begin{figure}[p]
  \centering
  \includegraphics[width=\linewidth]{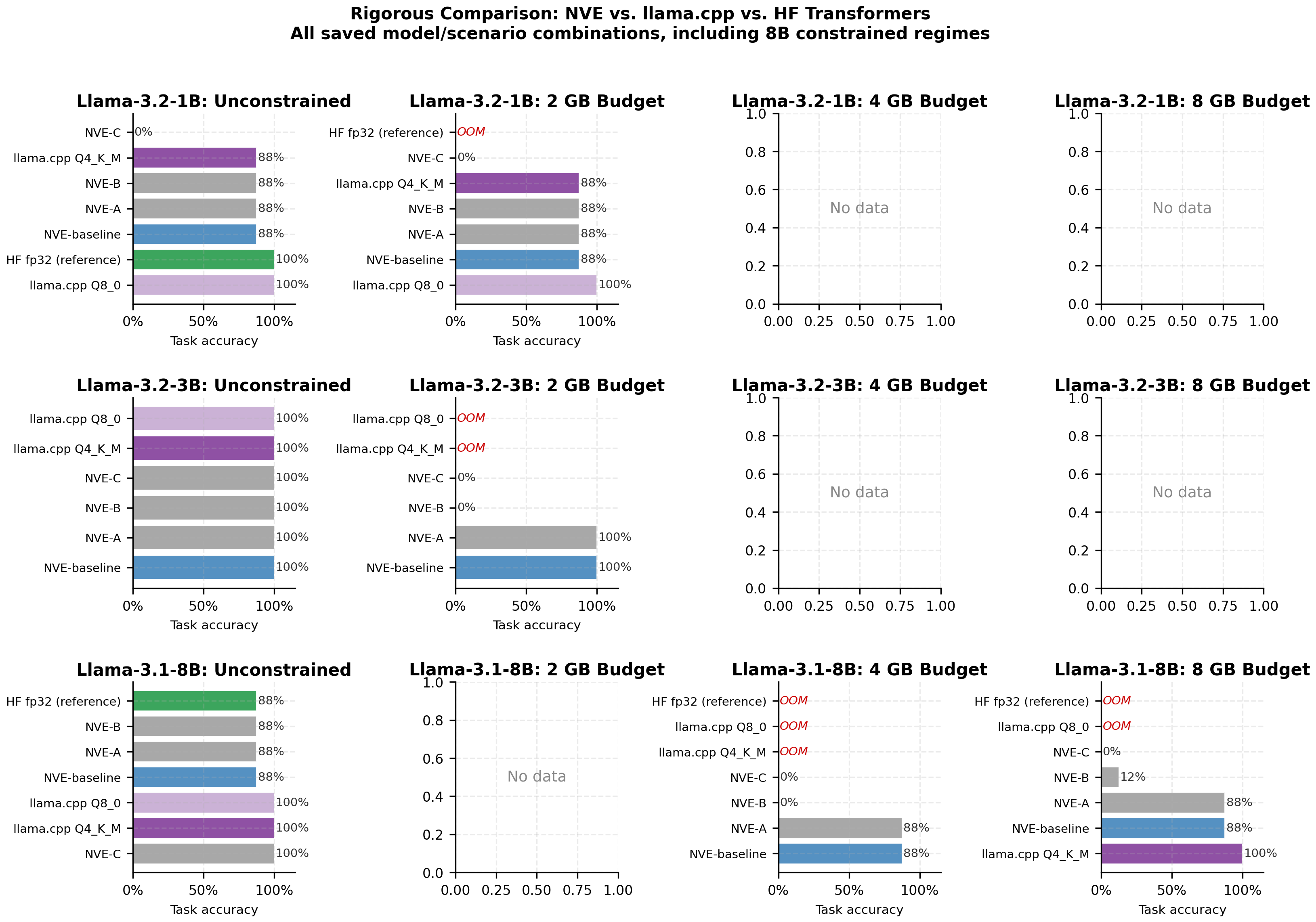}
  \caption{\textbf{Full system comparison: NVE vs.\ llama.cpp vs.\ HuggingFace.}
  Multi-panel comparison of task accuracy across saved model/budget scenarios, including
  8B unconstrained, 4\,GB, and 8\,GB settings. Hatched entries indicate OOM or unavailable
  baselines, highlighting that NVE's main advantage is preserving reachable operating
  points under tight memory budgets. Panels labelled ``No data'' (Llama-3.2-1B at 4/8\,GB,
  Llama-3.2-3B at 4/8\,GB, Llama-3.1-8B at 2\,GB) were not part of the sweep because the
  combination is either unconstrained for the model (1B/3B fit trivially at 4--8\,GB) or
  below the hot-only floor reported elsewhere (8B at 2\,GB).}
  \label{fig:rigorous_comparison}
\end{figure}

\paragraph{Full system comparison.} Figure~\ref{fig:rigorous_comparison} provides a
comprehensive side-by-side comparison of NVE, llama.cpp, and HuggingFace FP16 across
saved model-and-budget scenarios, including the larger 8B constrained regimes. Rather
than collapsing everything into a single scalar, the figure shows per-scenario task
accuracy and marks out-of-memory baselines explicitly. The relevant
comparison is scenario-dependent: not only ``how accurate is each system''
but also ``which operating points remain reachable at all.'' Viewed this way,
NVE's advantage is not only higher throughput under comfortable budgets but
access to tight 3B/8B memory regimes in which baseline systems do not run.

\paragraph{Representative failure examples.} The collapse of aggressive profile-guided
quantization is visible directly in the generated text, not only in summary metrics. In the
1B and constrained larger-model failures, the model falls into short repetitive loops or
template fragments rather than producing merely lower-quality answers. Table~\ref{tab:failure_examples}
provides representative examples from the raw outputs. The transition is a
sharp boundary between \emph{degraded but usable} and \emph{structurally
broken} operating points.

\begin{table}[H]
\centering
\caption{\textbf{Representative corrupted generations from unstable operating points.}
These examples are taken from saved raw outputs and illustrate the qualitative signature of
failure: repetitive boilerplate and malformed continuations rather than small semantic
mistakes.}
\label{tab:failure_examples}
\footnotesize
\setlength{\tabcolsep}{4pt}
\begin{tabular}{p{0.13\linewidth}p{0.17\linewidth}p{0.23\linewidth}p{0.37\linewidth}}
\toprule
Model & Scenario / config & Prompt prefix & Representative output \\
\midrule
Llama-3.2-1B & unconstrained / C & ``The three branches \ldots'' &
\texttt{the following. 1. The first branch of the first branch \ldots} \\
Llama-3.2-1B & unconstrained / C & \texttt{def fibonacci(n):} &
\texttt{return fibonacci(n) if fibonacci(n) is fibonacci(n) else fibonacci(n) \ldots} \\
Llama-3.2-1B & constrained 2\,GB / C & ``Photosynthesis is \ldots'' &
\texttt{are grown. The plants are grown. The plants are grown. \ldots} \\
Llama-3.1-8B & constrained 4\,GB / B,C & 8-task suite &
\textit{Task-level reports show 0\% accuracy despite elevated tok/s, indicating partially or fully incoherent output rather than a small quality drop.} \\
Llama-3.2-1B & 0.5 bpw / C & ``general relativity'' &
\texttt{the following is the case. 1. The following is the case. 2. The following is the case. 3. The following is the case \ldots} \\
Llama-3.2-1B & 1.0 bpw / C & ``general relativity'' &
\texttt{the following is the case. 1. The general relativity is the case. 2. The general relativity is the case \ldots} \\
Llama-3.2-1B & 1.5 bpw / C & ``general relativity'' &
\texttt{the following is the case. The following is the case. The following is the case \ldots} \\
\bottomrule
\end{tabular}
\end{table}

\paragraph{Failure signature is consistent across configurations.} The three new rows
above are from independent BPW-sweep runs at 0.5, 1.0, and 1.5 bits per weight, all
with Config~C. The failure signature, a short
template fragment repeating with minor surface variation, is remarkably stable across
bit-widths. This is useful for deployment monitoring: a degenerate repetition-rate
detector would catch Config~C collapse in a single generation, whereas a perplexity-only
metric might miss it (the repeated template is locally high-probability even as the
sequence is semantically broken).

\section{Proof of the MCAP Recovery Proposition}
\label{app:mcap_proof}

We restate the proposition for convenience. Let $s_i$ denote the population-level MCAP
score for layer $i \in \{1, \ldots, L\}$ and $\hat{s}_i^{(N)}$ its empirical estimate from
$N$ calibration prompts, computed as a sample mean over the $N$ per-prompt activation
statistics. We assume the $N$ per-prompt statistics behave as iid draws with a
sub-Gaussian envelope of parameter $\sigma_a$, so that the sample-mean deviation
satisfies
\[
  \Pr\!\left[ \left| \hat{s}_i^{(N)} - s_i \right| \geq t \right] \leq 2 \exp\!\left( -\frac{N t^2}{2 \sigma_a^2} \right).
\]
This is an \emph{idealization}: the 12 prompts are curated for topic coverage
(Section~\ref{sec:mcap}), not iid draws, so we treat the bound as a scaling argument
rather than an operational guarantee (see ``On tightness'' below).

Let $s_{(1)} \geq s_{(2)} \geq \cdots \geq s_{(L)}$ be the sorted population scores and
$\Delta_k := s_{(k)} - s_{(k+1)}$ the top-$k$ gap. Let $\mathcal{T}_k$ be the true top-$k$
set and $\hat{\mathcal{T}}_k^{(N)}$ the empirical one. Then:
\[
  \Pr\!\left[ \hat{\mathcal{T}}_k^{(N)} \neq \mathcal{T}_k \right] \leq 2\,k(L-k) \cdot \exp\!\left( -\frac{N \Delta_k^2}{8 \sigma_a^2} \right).
\]

\paragraph{Proof.} A recovery failure $\hat{\mathcal{T}}_k^{(N)} \neq \mathcal{T}_k$
requires at least one inversion: some pair $(i, j)$ with $i \in \mathcal{T}_k$,
$j \notin \mathcal{T}_k$, and $\hat{s}_i^{(N)} < \hat{s}_j^{(N)}$. For this pair,
$s_i - s_j \geq \Delta_k$ by definition of the top-$k$ gap, so the inversion requires
\[
  (\hat{s}_i^{(N)} - s_i) - (\hat{s}_j^{(N)} - s_j) \leq -(s_i - s_j) \leq -\Delta_k.
\]
By the triangle inequality, this forces at least one of $|\hat{s}_i^{(N)} - s_i|$ or
$|\hat{s}_j^{(N)} - s_j|$ to exceed $\Delta_k / 2$. The sub-Gaussian tail bounds each
such event by $2 \exp(-N \Delta_k^2 / 8 \sigma_a^2)$.

There are at most $k(L-k)$ straddling pairs, and each contributes two deviation events,
so a union bound gives
\[
  \Pr\!\left[ \hat{\mathcal{T}}_k^{(N)} \neq \mathcal{T}_k \right] \leq 2\,k(L-k) \cdot \exp\!\left( -\frac{N \Delta_k^2}{8 \sigma_a^2} \right).
\]
The commonly cited specialization to top-1 recovery ($k=1$) gives prefactor $2(L-1)$.
\hfill $\blacksquare$

\paragraph{Numerical instantiation for Llama-3.1-8B, top-1.} From
Table~\ref{tab:mcap_scores}, $L = 32$, $s_{(1)} = 262$, $s_{(2)} \approx 146$\footnote{%
Second-largest 8B score read from the saved profile JSON; Table~\ref{tab:mcap_scores}
reports only the full range (58--262).}
(second-largest final-block score), giving $\Delta_1 \approx 116$. The empirical
per-layer norm standard deviation across the 12 calibration prompts is
$\sigma_a \approx 64$ (measured directly from the saved profiling run). Then
\[
  \frac{N \Delta_1^2}{8 \sigma_a^2} = \frac{12 \cdot 116^2}{8 \cdot 64^2} \approx 4.9,
\]
so the top-1 bound $2(L-1)\,e^{-4.9} \approx 62 \cdot 0.0074 \approx 0.46$ is non-vacuous
but loose. For top-$k$ with larger $k$, the $k(L-k)$ prefactor grows and the raw union
bound becomes vacuous; tighter constants require assumptions we do not formally verify.
The observed split-half top-$k$ overlap of 1.0 (Table~\ref{tab:mcap_scores}) is
consistent with a much tighter true concentration; we report it as empirical evidence
that the operational behavior is well inside the regime the proposition describes
qualitatively.

\paragraph{On tightness.} The proposition gives a qualitatively correct dependency
structure ($N$ scales with $\sigma_a^2 / \Delta_k^2$) and a conservative constant. The
split-half overlap data suggests the true constant is smaller than the union bound
predicts, which is expected: Hoeffding-style bounds are famously loose for highly
concentrated empirical distributions like the per-layer norm statistics we observe.

\section{Reproducibility}
\label{app:reproducibility}

\paragraph{Code and data availability.}
The full source tree, calibration prompts, captured run logs, and paper
source are released under the MIT License at
\url{https://github.com/genovationtech/nve}. The layout relevant to the
claims in this paper is:

\begin{itemize}[leftmargin=*, itemsep=1pt]
\item \texttt{src/profiler.rs} --- MCAP profiler
  (Section~\ref{sec:mcap}): 12-prompt calibration, per-layer
  activation variance, importance scoring, 338-byte JSON profile emission.
\item \texttt{src/importance\_cache.rs} --- architecture-keyed profile
  cache referenced in Section~\ref{sec:system}.
\item \texttt{src/quantize.rs}, \texttt{src/cuda\_kernels.rs} ---
  W4A8 / W4A16 dispatch and CUDA kernels
  (Section~\ref{sec:w4a8_kernel}).
\item \texttt{src/decode\_graph.rs} --- CUDA graph-captured decode path
  (Section~\ref{sec:w4a8_kernel}, \S CUDA graph capture).
\item \texttt{src/pager.rs}, \texttt{src/tier.rs},
  \texttt{src/paged\_model.rs} --- three-tier (GPU / RAM / SSD) weight
  pager (Section~\ref{sec:paging}).
\item \texttt{src/arch.rs}, \texttt{src/generic\_model.rs} --- generic
  transformer front-end covering the eight evaluated architectures
  (GPT-2 through Qwen2-7.6B).
\item \texttt{src/cli/} --- \texttt{nve} binary entry point
  (profile / run / bench subcommands).
\item \texttt{benchmarks/}, \texttt{src/benchmark.rs},
  \texttt{src/abc\_benchmark.rs} --- throughput, PPL, HellaSwag, and
  8-task harness used for Sections~\ref{sec:throughput}
  and~\ref{sec:quality}.
\item \texttt{evidence/} --- dated capture directories
  (\texttt{2026-04-12} through \texttt{2026-04-18}) holding the Modal
  runner scripts and raw JSON logs for every figure and table,
  including the cross-silicon profile hash and threshold ablation.
\item \texttt{tests/}, \texttt{examples/} --- integration tests and
  minimal runnable examples (\texttt{examples/test\_real\_model.py},
  \texttt{examples/test\_tiered\_serving.py}).
\item \texttt{docs/} --- architecture notes, quantization details, and
  the streaming profiler design referenced in the main text.
\item \texttt{paper/mcap.tex} --- source of this document; the
  \texttt{figures/} subtree holds every plot included here.
\end{itemize}

Hugging Face access tokens in benchmark scripts are read from the
\texttt{HF\_TOKEN} environment variable; no credentials are committed to
the repository.

\begin{itemize}[leftmargin=*, itemsep=2pt]
\item \textbf{Hardware}: GPU experiments on Modal cloud (NVIDIA T4 sm\_75 /
  A10G sm\_86). CPU-only experiments on a commodity AMD cloud instance
  (2~vCPU, 3.8\,GB RAM).
\item \textbf{Software}: CUDA 12.1, driver 535.xx, Rust 1.78, Python 3.11.
  Each saved run records the Modal container image digest.
\item \textbf{Build}: CUDA kernels compiled with
  \texttt{nvcc -O3 --use\_fast\_math -arch=compute\_75} into a shared library
  linked to a release-mode Rust build with CUDA features enabled.
\item \textbf{Seeds}: decode uses greedy (argmax) sampling throughout; no RNG
  seed is needed for the tok/s, PPL, HellaSwag, or 8-task results. The 12
  calibration prompts are a fixed list held constant across runs, not sampled
  at run time.
\item \textbf{Determinism}: each saved run records the git commit SHA and
  the Modal image digest; the commit used to generate every figure in the
  paper is logged. The 338-byte cross-silicon profile referenced in
  Section~\ref{sec:memory} has SHA-256\\
  {\scriptsize\texttt{ac4eea60d1675b12bc83e6e7ac2bd6\allowbreak 752c3fda90262306a5985fe843ae8fd149}}.
\end{itemize}

\end{document}